\documentclass[10pt,twocolumn,letterpaper]{article}

\usepackage[accsupp]{axessibility} 
\usepackage[pagenumbers]{wacv} 

\usepackage{graphicx}
\usepackage{amsmath}
\usepackage{amssymb}
\usepackage{booktabs}

\usepackage[pagebackref,breaklinks,colorlinks]{hyperref}

\usepackage[capitalize]{cleveref}
\crefname{section}{Sec.}{Secs.}
\Crefname{section}{Section}{Sections}
\Crefname{table}{Table}{Tables}
\crefname{table}{Tab.}{Tabs.}

\usepackage{amsmath}
\usepackage{amssymb}
\usepackage{mathtools}
\usepackage{multirow}

\usepackage{algorithmic}
\usepackage{algorithm}
\usepackage{tablefootnote}
\usepackage{threeparttable}

\newcommand{\x}{\mathbf{x}}
\newcommand{\z}{\mathbf{z}}
\renewcommand{\a}{\mathbf{a}}
\renewcommand{\b}{\mathbf{b}}
\newcommand{\w}{\mathbf{w}}
\newcommand{\zero}{\mathbf{0}}
\newcommand{\Id}{\mathbf{I}}
\newcommand{\R}{\mathbb{R}}
\newcommand{\Gauss}[2]{\mathcal{N}\left(#1,#2\right)}

\newcommand{\lt}{\ensuremath <}

\allowdisplaybreaks
\def\wacvPaperID{1535} 
\def\confName{WACV}
\def\confYear{2025}

\begin{document}

\title{ARD-VAE: A Statistical Formulation to Find the Relevant Latent Dimensions of Variational Autoencoders}

\author{Surojit Saha\\
The University of Utah, USA\\
{\tt\small surojit.saha@utah.edu}
\and
Sarang Joshi\\
The University of Utah, USA \\
{\tt\small sarang.joshi@utah.edu}
\and
Ross Whitaker\\
The University of Utah, USA \\
{\tt\small whitaker@cs.utah.edu}
}
\maketitle

\begin{abstract}
The variational autoencoder (VAE) \cite{rezende2014stochastic,kingma2014auto} is a popular, deep, latent-variable model (DLVM) due to its simple yet effective formulation for modeling the data distribution. Moreover, optimizing the VAE objective function is more manageable than other DLVMs. The bottleneck dimension of the VAE is a crucial design choice, and it has strong ramifications for the model's performance, such as finding the hidden explanatory factors of a dataset using the representations learned by the VAE. However, the size of the latent dimension of the VAE is often treated as a hyperparameter estimated empirically through trial and error. To this end, we propose a statistical formulation to discover the relevant latent factors required for modeling a dataset. In this work, we use a \emph{hierarchical} prior in the latent space that estimates the variance of the latent axes using the encoded data, which identifies the relevant latent dimensions. For this, we replace the fixed prior in the VAE objective function with a hierarchical prior, keeping the remainder of the formulation unchanged. We call the proposed method the \emph{automatic relevancy detection in the variational autoencoder} (ARD-VAE)\footnote{\url{https://github.com/Surojit-Utah/ARD-VAE}}. We demonstrate the efficacy of the ARD-VAE on multiple benchmark datasets in finding the relevant latent dimensions and their effect on different evaluation metrics, such as FID score and disentanglement analysis. 
\end{abstract}

\section{Introduction}
\label{sec:intro}
Deep latent variable models (DLVMs) have emerged as powerful tools for modeling data distributions, generating novel examples from a distribution, and learning useful encodings for downstream applications. Many deep latent variable models have been proposed, including the variational autoencoder (VAE), adversarial autoencoder (AAE) \cite{makhzani2016adversarial}, Wasserstein autoencoder(WAE) \cite{tolstikhin2017wasserstein}, and diffusion-based models \cite{score_based_diffusion,image_gen_diffusion}. Besides the unsupervised representation learning \cite{kingma2014auto,makhzani2016adversarial}, the latent mappings produced by DLVMs find potential application in semi-supervised \cite{kingma2014auto,vae_semi_sup_2019}, weakly supervised \cite{vae_weak_sup_2020,vae_weak_sup_2023} and few-shot learning \cite{vae_few_shot_2019,vae_few_shot_2020,vae_few_shot_2023}. Moreover, the quality of the generated samples produced by the diffusion probabilistic models \cite{score_based_diffusion,image_gen_diffusion} is comparable to state-of-the-art (SOTA) GANs \cite{Style_GAN_2021}. All these properties make DLVMs an important and active area of research.

Among DLVM approaches, the VAE \cite{rezende2014stochastic,kingma2014auto} has a strong theoretical foundation that uses \emph{variational inference} to approximate the true posterior by a surrogate distribution. The VAE uses a \emph{inference/recognition} model to estimate the parameters of the surrogate posterior distribution in an amortized framework (shared parameters). The use of the reparameterization trick in VAEs leads to efficient optimization of the lower bound on the data log-likelihood. In addition, it is simple to train a neural network with the VAE objective function relative to the AAE \cite{makhzani2016adversarial} and WAE \cite{tolstikhin2017wasserstein}(due to the use of the discriminators in these models). These advantages have made the VAE a popular DLVM, which has even helped in progressing basic science research \cite{vae_basic_syn_mol_2018,vae_basic_mol_struct_2023,vae_basic_rna_seq_2023}. However, the VAE often fails to match the aggregate distributions in the latent space (manifested by \emph{pockets/holes} in the encoded distribution) \cite{rosca2018distribution,dai2019_2sVAE} and suffers from posterior collapse (uninformative latent variables) \cite{ELBO_surgery_2016,posterior_collapse_ICLR_2019,posterior_collapse_Neurips_2019}. Several methods have been proposed for better matching of marginal posteriors \cite{Factor_VAE_ICML_2018,TC_VAE_Neurips_2019,gens_saha_2022} and to alleviate posterior collapse \cite{posterior_collapse_VQ_VAE_2017_NIPS,posterior_collapse_delta_VAE_ICLR_2019}

In DLVMs, we generally assume that the observed data existing in a high dimensional space $(\R{}^{D})$, such as images, can be represented by a low dimensional manifold $(\R{}^{M})$ embedded in that space, where $M \lt \lt D$. The size of the manifold, aka \emph{intrinsic} data dimension, can be interpreted as the number of latent factors used to produce the observed data (ignoring the noise). The bottleneck size of DLVMs represents the \emph{intrinsic} data dimension. However, the true \emph{intrinsic} dimension is generally unknown for real-world datasets, and finding the correct dimension remains an open and challenging problem. 
Studies have shown that incorrect bottleneck dimensions of DLVMs result in inaccurate modeling of the data distribution, which subsequently affects the interpretation of the latent representations produced by a DLVM \cite{tolstikhin2017wasserstein,dai2019_2sVAE}. 
However, only a few methods \cite{vae_prune_mask_AAE_reg_2020,vae_prune_L0_reg_2021} have addressed this problem in DLVMs. For instance, the MaskAAE \cite{vae_prune_mask_AAE_reg_2020} introduces a trainable masking vector in the latent space of a deterministic autoencoder (used in \cite{makhzani2016adversarial,tolstikhin2017wasserstein}) to determine the number of active dimensions. In \cite{vae_prune_L0_reg_2021}, the authors apply a $L_0$ regularization on the masking vector to prune unnecessary dimensions in the VAE. In both methods, additional regularization loss is added to the primary objective function with associated hyperparameters. It is observed that the proposed methods are sensitive to the choice of the neural architecture architectures and settings of the hyperparameters used in the objective function. Thus, the methods are useful only with certain architectures or training strategies and do not generalize well to different scenarios.

Considering the importance of the bottleneck dimension in the VAE and limitations in the existing methods, we propose a method to determine the relevant latent dimensions for a dataset using a \emph{hierarchical} prior  \cite{Bishop_Bayesian_PCA_1998,Bishop_Bayesian_PCA_1999,rvm_tipping_2001} introduced in \cite{Neal_ARD_1996}, where the prior distribution, $p(\z{})$, is dependent on another random variable, $\boldsymbol\alpha$, a \emph{hyperprior}. In the proposed statistical formulation, the hierarchical prior distribution is defined as $p(\z{} \mid \boldsymbol\alpha)$, where $\boldsymbol\alpha$ estimates the precision of the latent dimensions using encoded data. The variance ($\boldsymbol\alpha^{-1}$) along the latent axes is used to determine the relevant latent factors required to represent a dataset, such that the unwanted latent variables, through training, obtain very low variance. Thus, the ARD-VAE introduces a natural extension to the fixed prior used in the VAE that is flexible and is learned from the data. Unlike other approaches \cite{vae_prune_mask_AAE_reg_2020,vae_prune_L0_reg_2021}, we do not modify the ELBO objective of the VAE, except the use of the hierarchical prior. Following are the contributions of this work:
\begin{itemize}
  \item We propose the ARD-VAE to discover relevant axes in the VAE using a hierarchical prior without modification of the ELBO in the VAE.
  \item The ARD-VAE is insensitive to the choice of the autoencoder architecture and optimization strategies.
  \item Empirical evaluations demonstrate the effectiveness of the ARD-VAE in modeling data distributions and finding relevant generative factors.
\end{itemize}

\section{Related Work}\label{sec:relaed_wroks}
VAEs often fail to match the aggregate posterior that subsequently affects the modeling of the data distribution \cite{rosca2018distribution,dai2019_2sVAE,avae_saha_2023}. Matching the aggregate posterior also makes the latent dimensions independent, thus encouraging disentanglement of the latent factors \cite{DIP_VAE_ICLR_2018,Factor_VAE_ICML_2018,TC_VAE_Neurips_2019}. The regularized autoencoder (RAE) \cite{RAE} estimates the aggregate encoded distribution (using an ex-post density estimator) to model the data distribution, where it replaces the stochastic autoencoders used in VAEs with its deterministic counterpart. 
All these analyses encourage the matching of aggregate distribution in VAEs as done in other DLVMs, such as the AAE \cite{makhzani2016adversarial} and WAE \cite{tolstikhin2017wasserstein}, to improve the performance of the model. In this work, we study the dimensionality of the original VAE formulation (which matches conditional distributions to the prior), with the understanding that it applies readily to aggregate-matching methods.

The mismatch between the size of the bottleneck dimension (chosen in VAEs) and the intrinsic data dimension is shown to be a critical factor in designing a VAE. A detailed analysis of the effect of mismatch between the dimensions is done in \cite{dai2019_2sVAE}, which elucidates the importance of finding the correct size of the latent space. Finding the intrinsic data dimension is important for any DLVM \cite{tolstikhin2017wasserstein} and is not unique to VAEs. 

A few methods have been proposed in the recent past to address this issue; one is designed for VAEs \cite{vae_prune_L0_reg_2021} and the other one for AAEs \cite{vae_prune_mask_AAE_reg_2020}.
The method in \cite{vae_prune_L0_reg_2021} (referred to herein as GECO-$L_0$-ARM-VAE) introduces a boolean gating vector (matching the bottleneck dimension) to find the cardinal latent axes required for modeling a dataset in VAEs. The latent encodings are combined with the shared gating vector (using the element-wise product) to optimize the ELBO objective function of the VAE. To encourage sparsity in the gating vector, the authors have augmented the ELBO with a $L_0$ regularization loss on the gating vector. However, optimization of the $L_0$ loss using gradient descent is challenging, and it uses the $L_0$-ARM \cite{L0_ARM_L0_reg_2019} to update the gradients. To make the effect of the regularization of the latent representations more interpretable to the users, they use a threshold for the reconstruction loss, as in \cite{GECO_2018}. In optimizing the objective function of the  GECO-$L_0$-ARM-VAE, the parameters of the gating vector are updated after the desired reconstruction loss is achieved, i.e., after the model parameters have learned sufficient statistics of the data. After that, all model parameters are trained jointly.

The MaskAAE \cite{vae_prune_mask_AAE_reg_2020} uses a trainable masking vector similar to \cite{vae_prune_L0_reg_2021} to determine the relevant latent dimensions. Unlike the method in \cite{vae_prune_L0_reg_2021}, the MaskAAE is designed for the AAE \cite{makhzani2016adversarial} and WAE \cite{tolstikhin2017wasserstein} that uses deterministic autoencoders and matches the aggregate (or marginal) posterior to the prior. The Hadamard products of the latent encodings with the global masking vector are used as inputs to the decoder and discriminator. The masking vector is initialized with continuous values for gradient updates. A regularization loss, the masking loss, is added to the objective function of the AAE, along with an additional loss term to the reconstruction loss. 

Both the GECO-$L_0$-ARM-VAE and MaskAAE use a lot of hyperparameters and different initialization strategies, which are most likely to be sensitive to neural architectures and other optimization settings. The MaskAAE uses a complex training recipe to optimize different loss functions and update hyperparameters. Thus, the typical settings used in both methods raise questions about using the methods on new datasets, which might require changes in the VAE/AAE architecture. In contrast, the proposed method has a single hyperparameter to balance the reconstruction and regularization loss in the objective function (similar to any DLVM). Furthermore, it is flexible enough to be used with virtually any VAE architecture and, thus, can be trained on any new dataset.

\section{Automatic Relevancy Detection in the Variational Autoencoder}
\subsection{Background} The VAE is a generative model that approximates the unknown data distribution, $p(\x{})$, by learning the joint distribution of the latent variables, $\z{}, \text{where } \z{} \in \R{}^{L}$, and the observed variables, $\x{}, \text{where } \x{} \in \R{}^{D}$. The VAE maximizes the lower bound on the data log likelihood in \cref{eq:VAE_ELBO}, known as the evidence lower bound (ELBO),
\begin{align}\label{eq:VAE_ELBO}
\max_{\theta, \phi} \mathbb{E}_{p(\x{})} \Bigl[ \mathbb{E}_{q_{\phi}(\z{} \mid \x{})} \log p_{\theta}(\x{} \mid \z{}) - \operatorname{KL} \Bigl( q_{\phi}(\z{} \mid \x{}) \lvert\rvert p(\z{}) \Bigr) \Bigr].
\end{align}


VAEs use a probabilistic encoder, $\mathbf{E_{\phi}}$, and a probabilistic decoder, $\mathbf{D_{\theta}}$, to represent $q_{\phi}(\z{} \mid \x{})$ and, $p_{\theta}(\x{} \mid \z{})$, respectively. Both $\mathbf{E_{\phi}}$ and $\mathbf{D_{\theta}}$ are usually deep neural networks parameterized by $\phi$ and $\theta$, respectively. The prior distribution,  $p(\z{})$, is chosen to be $\Gauss{\zero}{\Id}$ and
the surrogate posterior is a Gaussian distribution with diagonal covariance (under the assumption of independent latent dimensions), which is defined as follows:
\begin{align}\label{eq:VAE_Surr_posterior}
q_{\phi}(\z{} \mid \x{}) = \Gauss{\boldsymbol\mu_{\x{}}}{\boldsymbol\sigma_{\x{}}^2\Id}, \text{where } \boldsymbol\mu_{\x{}}, \boldsymbol\sigma_{\x{}}^2 \leftarrow \mathbf{E_{\phi}}(\x{}).
\end{align}
The choice of the Gaussian distribution as the posterior, $q_{\phi}(\z{} \mid \x{})$, helps in efficient computation of the $\mathbb{E}_{q_{\phi}(\z{} \mid \x{})} p_{\theta}(\x{} \mid \z{})$ using the reparameterization trick introduced by the VAE,
\begin{align}\label{eq:VAE_reparam}
\z{} \leftarrow \boldsymbol\mu_{\x{}} + \epsilon \odot \boldsymbol\sigma_{\x{}}, \text{where } \epsilon \sim \Gauss{\zero}{\Id} 
\end{align}
Moreover, we have a closed form solution for the $\operatorname{KL}$ divergence between Gaussian distributions in \cref{eq:VAE_ELBO}. The multivariate Gaussian distribution or Bernoulli distribution is used as the likelihood distribution, $p_{\theta}(\x{} \mid \z{})$. The VAE jointly optimizes the parameters of the $\mathbf{E_{\phi}}$ and $\mathbf{D_{\theta}}$ using the objective function in \cref{eq:VAE_ELBO}, under the modeling assumptions discussed above.

\subsection{Formulation for Relevance-Aware Prior}
In this paper, we propose a statistical formulation to determine the number of relevant latent axes for a dataset using a given encoder-decoder architecture. To this end, we allow the prior distribution, $p(\z{})$ (a Gaussian), to have different variance along each latent axis, represented by another random variable $\boldsymbol\alpha$, which forms a hyperprior. Under this setting, the latent axes with high variances should represent important latent factors and unnecessary latent dimensions are expected to exhibit relatively low variances. To this end, we use a hierarchical prior \cite{Neal_ARD_1996,Bishop_Bayesian_PCA_1998,Bishop_Bayesian_PCA_1999} on independent latent dimensions (as in VAEs) defined as,
\begin{align}\label{eq:ARD-VAE_prior}
p(\z{} \mid \boldsymbol\alpha) &= \prod_{l=1}^{L} \mathcal{N}\left(z_{l}; 0, \alpha_{l}^{-1} \right), \\
p(\boldsymbol\alpha) &= \prod_{l=1}^{L} \text{Gamma}(\alpha_{l} ; a_l^0, b_l^0).
\end{align}
where $\alpha_{l}^{-1}$ is the variance along latent axis $l$. The Gamma distribution is used to model the distribution over the independent precision variable ($\alpha_{l}$) as it is a suitable choice for modeling scale parameters (variance of the Gaussian distribution) \cite{Berger_Statistics_Gamma}. Moreover, the Gamma distribution is a conjugate prior to the $p(\z{} \mid \boldsymbol\alpha)$ (tractable posterior) \cite{Bishop_PRML,rvm_tipping_2001}. On integrating the hyperparameter, $\boldsymbol\alpha$, we get the marginalized prior distribution, $p(\z{})$, that is a Student's $t$-distribution:
\begin{align}
p(\z{}) &= \int p(\z{} \mid \boldsymbol\alpha) p(\boldsymbol\alpha) d\boldsymbol\alpha \label{eq:ARD-VAE_prior_marg_wo_data} \\
p(\z{}) &= \prod_{l=1}^{L} {\frac {\ b_{l}^{a_{l}}\Gamma (a_{l}+\frac{1}{2})\ }{\ {\sqrt {2\pi \ }}\ \Gamma (a_{l})}}\left(\ b_{l}+{\frac {z_{l}^{2}\ }{2}}\ \right)^{-(a_{l} + \frac{1}{2})}. \label{eq:Student_t_wo_data}
\end{align}
Considering ${a}_{l}=0$ and ${b}_{l}=0$ (uniform distribution over the parameters of the Gamma distribution), the probability over the latent variables, $p(\z{})=\prod_{l=1}^{L} p(z{}_{l})$, will be peaked around zero, as $p(z{}_{l}) \propto 1/{\mid z{}_{l}\mid}$ \cite{rvm_tipping_2001,Bishop_PRML}. This property of the prior distribution encourages \emph{sparsity} that is used in the proposed method to discover the relevant latent axes.

The parameters of the Gamma distribution are estimated analytically using data in the latent space, $D{}_{\z{}}$, produced using the posterior distribution of the VAE estimated by the encoder, $\mathbf{E_{\phi}}$ (refer to algorithm 2 in the supp). Thus, the distribution over $\boldsymbol\alpha$ conditioned on data, $D{}_{\z{}}$, is defined as $p(\boldsymbol\alpha \mid D{}_{\z{}})$. In this limited context, $p(\boldsymbol\alpha \mid D{}_{\z{}})$ can be interpreted as a posterior distribution, which is $ \propto p(D{}_{\z{}} \mid \boldsymbol\alpha) p(\boldsymbol\alpha)$. Given $p(\boldsymbol\alpha)$ as the Gamma distribution (prior), the posterior $p(\boldsymbol\alpha \mid D{}_{\z{}})$ is also a Gamma distribution (conjugate prior) when $p(D{}_{\z{}} \mid \boldsymbol\alpha)$ is a Gaussian distribution (likelihood). The parameters of $p(\boldsymbol\alpha \mid D{}_{\z{}})$ are derived analytically for each latent axis independently. The parameters of $p(\alpha_{l} \mid D{}_{\z{}})$ for each axis for a given mean $\boldsymbol{\mu^{\boldsymbol\alpha}} \in \R{}^{L}$ (mean of the likelihood distribution, $p(D{}_{\z{}} \mid \boldsymbol\alpha)$) are \cite{Baeyesian_conjugate}:
\begin{align}
a_{l} &= a_{l}^{0} + \frac{n}{2}, \label{eq:Gamma_a} \\
b_{l} &= b_{l}^{0} + \frac{\sum_{i}^{n}({z}_{l}^i - \mu^{\boldsymbol\alpha}_{l})^{2}}{2} \label{eq:Gamma_b}, 
\end{align}\label{eq:HP_update}
where $n$ is the number of samples in $D{}_{\z{}}$ and $\mathbf{z}^i \in D{}_{\z{}}$. The parameters of the Gamma distribution for all the latent axes are denoted by $\boldsymbol{a}_{L} \text { and } \boldsymbol{b}_{L}$.

With the introduction of data, $D{}_{\z{}}$, the hierarchical prior distribution in \cref{eq:ARD-VAE_prior} ($p(\z{} \mid \boldsymbol\alpha)$) is modified to $p(\z{} \mid \boldsymbol\alpha, D{}_{\z{}})$. The new distribution can be factorized as $p(\z{} \mid \boldsymbol\alpha, D{}_{\z{}})=p(\z{} \mid \boldsymbol\alpha) p(\boldsymbol\alpha \mid D{}_{\z{}})$. For optimization of the ELBO objective in \cref{eq:VAE_ELBO}, we need to remove $\boldsymbol\alpha$ from $p(\z{} \mid \boldsymbol\alpha, D{}_{\z{}})$. The hyperprior $\boldsymbol\alpha$ is removed by marginalization (as in \cref{eq:ARD-VAE_prior_marg}), and that integration results in a Student's $t$-distribution for each latent variable, as shown in \cref{eq:Student_t} (please refer to exercise 2.46 of \cite{Bishop_PRML}). The Student's $t$-distribution encourages sparsity in the latent variable $\z{}$ (refer to \cref{eq:Student_t_wo_data} and section 5.1 in \cite{rvm_tipping_2001}). The degrees of freedom, $\nu_l$ (\cref{eq:Student_t_dof}), and $t$-statistics, $t_l$ (\cref{eq:Student_t_stat}), of the $t$-distributions are linear functions of the parameters of the Gamma distribution used for modeling $\boldsymbol\alpha$ (defined in \cref{eq:Gamma_a} and \cref{eq:Gamma_b}).
\begin{align}
p(\z{} \mid D{}_{\z{}}) &= \int p(\z{} \mid \boldsymbol\alpha) p(\boldsymbol\alpha \mid D{}_{\z{}}) d\boldsymbol\alpha \label{eq:ARD-VAE_prior_marg} \\
p(\z{} \mid D{}_{\z{}}) &= \prod_{l=1}^{L} \frac{1}{s_l}{\frac {\ \Gamma ({\frac {\ \nu_{l} +1\ }{2}})\ }{\ {\sqrt {\pi \ \nu_{l} \ }}\ \Gamma ({\frac {\nu_{l} }{2}})}}\left(\ 1+{\frac {t_{l}^{2}\ }{\nu_{l} }}\ \right)^{-(\nu_{l} +1)/2}, \label{eq:Student_t} \\ 
\text{where, } \nu_{l} &= 2*a_{l} \label{eq:Student_t_dof} \\
t_{l}^{2} &= \frac{(z_{l} - \mu_{l})^{2}}{b_{l}/a_{l}} \label{eq:Student_t_stat} \\
s_{l} &= \sqrt{b_{l}/a_{l} }. \\
\end{align}
With the marginalized distribution, $p(\z{} \mid D{}_{\z{}})$ (\emph{target distribution} in VAEs), we can optimize the KL divergence term in the ELBO objective function in \cref{eq:VAE_ELBO}. Thus, the objective function of the ARD-VAE is defined as follows:
\begin{align}\label{eq:ARD-VAE_ELBO}
\max_{\theta, \phi} \mathbb{E}_{p(\x{})} \Bigl[ &\mathbb{E}_{q_{\phi}(\z{} \mid \x{})} \log p_{\theta}(\x{} \mid \z{}) - \\ \nonumber
&\operatorname{KL} \Bigl( q_{\phi}(\z{} \mid \x{}) \lvert\rvert p(\z{} \mid D{}_{\z{}}) \Bigr) \Bigr].
\end{align}

The objective function of the proposed method did not modify the ELBO derivation used in the conventional VAE \cite{rezende2014stochastic}, except for the use of the hierarchical prior distribution. Unlike methods in \cite{vae_prune_mask_AAE_reg_2020,vae_prune_L0_reg_2021}, we do not add a regularizer to the objective function in \cref{eq:VAE_ELBO} to determine the relevant axes. The number of samples in the set $D{}_{\z{}}$ gives us the degrees of freedom, $\nu$, of the Student's $t$-distribution in \cref{eq:Student_t}. The $t$-distribution is approximately the Gaussian distribution for higher $\nu$\cite{Bishop_PRML,kachigan1982multivariate,muirhead1982aspects}, which allows for a more efficient, analytical expression for the KL-divergence as in \cref{eq:ARD-VAE_Gaussian_approx_prior}. Thus, the proposed formulation supports optimization of the ELBO under different scenarios, depending on the size of $D{}_{\z{}}$.

\begin{figure*}
  \begin{subfigure}{0.67\linewidth}
    \centerline{\includegraphics[width=0.95\textwidth]{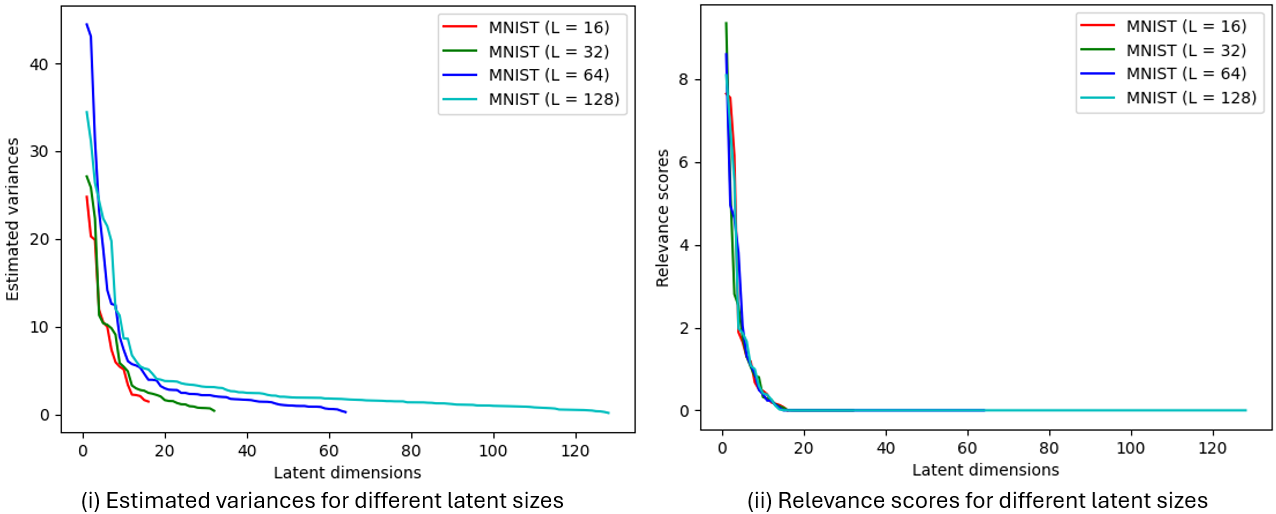}}
    \caption{(i) The estimated variances on the MNIST dataset using \ref{eq:ARD-VAE_estimated_var} for different latent sizes. (ii) The relevance scores computed using \ref{eq:ARD-VAE_wt_rel_axes_estimation} (on the same models) is used for estimating the \textsc{Active} dimensions.}
    \label{fig:MNIST_analysis}
  \end{subfigure}
  \begin{subfigure}{0.31\linewidth}
    \centerline{\includegraphics[width=1.05\textwidth]{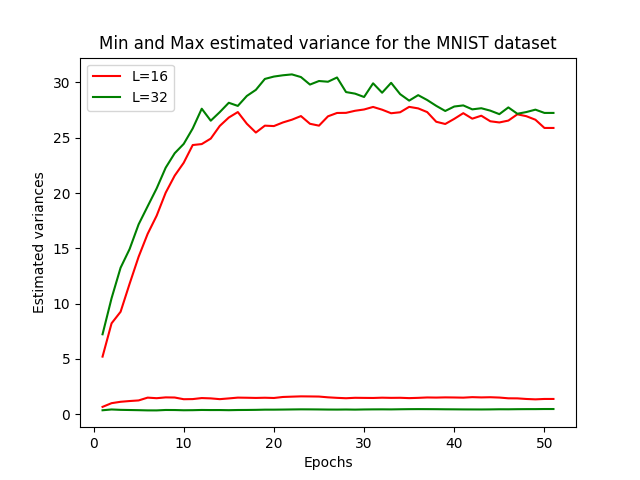}}
    \caption{The minimum and maximum variances estimated by the ARD-VAE while training on the MNIST dataset for multiple latent dimensions.}
    \label{fig:spread_est_var}
  \end{subfigure}
  \caption{(a) Relevance scores and (b) spread (min-max) of the estimated variances on the MNIST dataset while training the ARD-VAE.}
\end{figure*}

\subsection{Training the ARD-VAE}
In addition to the reconstruction loss, the objective function of the ARD-VAE (defined in \cref{eq:ARD-VAE_ELBO}) has a KL-divergence loss for matching the conditional posterior, $q_{\phi}(\z{} \mid \x{})$, to the target, $p(\z{} \mid D{}_{\z{}})$. The parameters of the target distribution, $p(\z{} \mid D{}_{\z{}})$ (in \cref{eq:Student_t_dof} and \cref{eq:Student_t_stat}), are dependent on the parameters of $p(\boldsymbol\alpha \mid D{}_{\z{}})$ (in \cref{eq:Gamma_a} and \cref{eq:Gamma_b}), which are learned from data. Thus, the ARD-VAE requires sample data to optimize the objective function in \cref{eq:ARD-VAE_ELBO} and learn the parameters of the hyperprior distribution (in \cref{eq:Gamma_a} and \cref{eq:Gamma_b}). To address this issue, the training data, $\mathcal{X{}}_{train}$ is split into $\mathcal{X{}}_{sgd}$ and $\mathcal{X{}}_{\boldsymbol\alpha}$, such that $\mid\mathcal{X{}}_{\boldsymbol\alpha}\mid \lt \mid\mathcal{X{}}_{sgd}\mid$. The set $\mathcal{X{}}_{sgd}$ is used to produce minibatches for optimization of the ARD-VAE objective function in \cref{eq:ARD-VAE_ELBO}. Samples in $\mathcal{X{}}_{\boldsymbol\alpha}$ are used to estimate the parameters, $\a{}_{L}$ and $\b{}_{L}$, of $p(\boldsymbol\alpha \mid D{}_{\z{}})$ using the data $D{}_{\z{}}$ produced by algorithm 2 (refer to the supp). 

The update of the dataset, $D{}_{\z{}}$, is set to lag for a few minibatches, as its update in every minibatch (an ideal scenario) would change the prior distribution in the KL divergence term in \cref{eq:ARD-VAE_ELBO}, leading to optimization challenges. The lagging $D{}_{\z{}}$ gives stability to the optimization of the ARD-VAE objective function. Moreover, the delayed update of $D{}_{\z{}}$ offers computational benefits. Using samples in the $D{}_{\z{}}$, the parameters of $p(\boldsymbol\alpha \mid D{}_{\z{}})$ are updated using \cref{eq:Gamma_a} and \cref{eq:Gamma_b}. In our work, we set $\boldsymbol{a}_{L}^{0}=0$ and $\boldsymbol{b}_{L}^{0}=0$ (uninformative) as in \cite{rvm_tipping_2001}. The mean vector $\boldsymbol{\mu^{\boldsymbol\alpha}}$ used in \cref{eq:Gamma_b} is also set to $\zero$. Under this configuration, $\boldsymbol{a}_{L}^{0}$ is fixed (samples size of $D{}_{\z{}}$) and only $\boldsymbol{b}_{L}^{0}$ is updated using the latest $D{}_{\z{}}$.

VAEs are trained in an unsupervised framework, having access to a large amount of data. Thus, the prior distribution can be approximated as a Gaussian distribution using sufficient samples in $D{}_{\z{}}$ (higher degree of freedom in the $t$-distribution) \cite{Bishop_PRML,kachigan1982multivariate,muirhead1982aspects}. With $p(\z{} \mid D{}_{\z{}})$ as an approximate Gaussian distribution, the KL divergence in \cref{eq:ARD-VAE_ELBO} is
\begin{align}\label{eq:ARD-VAE_Gaussian_approx_prior}
    &-\frac{L}{2} - \frac{1}{2}{\sum_{i=1}^{L}\log \sigma_i^2} + \frac{1}{2}{\sum_{i=1}^{L}\log {\hat{\sigma}_i}^2} + \frac{1}{2}\sum_{i=1}^{L} \frac{\mu_i^2 +  \sigma_i^2}{{\hat{\sigma}_i}^2}, \\ \nonumber
    &\text{where } \boldsymbol\mu, \boldsymbol\sigma^2 \leftarrow \mathbf{E_{\phi}}(\x{}), \x{} \in \mathcal{X{}}_{sgd} \text{ and } \\ \nonumber
    & \boldsymbol{\hat{\sigma}}^2 = \boldsymbol{b}_{L}/\boldsymbol{a}_{L}, \boldsymbol{a}_{L} \text { and } \boldsymbol{b}_{L} \text{ are estimated using the latest } D{}_{\z{}}.
\end{align}

In this work, we use the closed-form expression of the $\operatorname{KL}$ divergence in \cref{eq:ARD-VAE_ELBO}. However, one can still use the Student's $t$-distributions in \cref{eq:Student_t} as the prior distribution. In that scenario, one would use samples from $q_{\phi}(\z{} \mid \x{})$ to compute the KL divergence loss in \cref{eq:ARD-VAE_ELBO}.

A hyperparameter $\beta$ is used to balance the loss terms in the ARD-VAE objective function, as in \cite{beta_VAE_ICLR_2017, TC_VAE_Neurips_2019, Factor_VAE_ICML_2018, RAE}. We tune $\beta$ to achieve the desired reconstruction loss on the validation data. The outline of training the ARD-VAE is detailed in algorithm 1 in the supp.
For practical purposes, $D{}_{\z{}}$ is updated after every epoch ($u_{D{}_{\z{}}}=1$ in algorithm 1 in the supp) with a new subset of training samples, $\mathcal{X{}}_{\boldsymbol\alpha}$. Therefore, the SGD samples, $\mathcal{X{}}_{sgd}= \mathcal{X{}}_{train} - \mathcal{X{}}_{\boldsymbol\alpha}$, are also updated in the process. The typical size of $\mid\mathcal{X{}}_{\boldsymbol\alpha}\mid=10K$ for all the datasets studied in this work (refer to the ablation studies in the supp).

Unlike the VAE, the prior distribution $p(\z{} \mid D{}_{\z{}})$ of the ARD-VAE is not fixed and is updated during the training with data in the encoded space, $D{}_{\z{}}$. In \cref{fig:spread_est_var}, we track the minimum and maximum variance estimated using \cref{eq:ARD-VAE_estimated_var} while training the ARD-VAE on the MNIST dataset. The collapsed dimension indicated by the minimum variance has relatively low values across epochs. The variance of the relevant axis is way higher than the collapsed dimension and achieves stability after $\sim20$ epochs.

\subsection{Determining the Relevant Axes}
The ARD-VAE does not use a trainable masking vector as in \cite{vae_prune_L0_reg_2021,vae_prune_mask_AAE_reg_2020} to determine the relevant axes. Instead, it relies on the precision of the latent axes, $\boldsymbol\alpha$, to identify the spurious dimensions after the model is trained to convergence. The estimated variance is computed using the parameters, $\boldsymbol{a}_{L} \text { and } \boldsymbol{b}_{L}$, of the hyperprior distribution $p(\boldsymbol\alpha \mid D{}_{\z{}})$,
\begin{align}\label{eq:ARD-VAE_estimated_var}
    \boldsymbol{\hat{\sigma}}^2 = \boldsymbol{b}_{L}/\boldsymbol{a}_{L}.
\end{align}
As shown in \cref{fig:MNIST_analysis}(i), the estimated variance for the collapsed dimensions (no effect on the decoder output) has relatively low but non-zero values. Moreover, we observe variations in the estimates across latent dimensions in \cref{fig:MNIST_analysis}(i). Thus, the choice of a threshold or the percentage of variance (used in PCA) for selecting relevant dimensions that work under different scenarios is non-trivial.

For axes that are not relevant to the reconstruction, we have observed that the decoder produces virtually no variability in output in response to deviations along these axes. This is consistent with the fact that these axes are not being used for reconstruction. This motivates us to consider the deviation of the output, $\hat{\x{}} \in \R^{D}$ (produced by the decoder $\mathbf{D_{\theta}}$),  with respect to the mean representation, $\boldsymbol\mu_{\x{}} \in \R^{L} (\boldsymbol\mu_{\x{}} \leftarrow \mathbf{E_{\phi}}(\x{}))$, as the measure of relevance. We leverage this idea to find a weight vector $\w{}_{\boldsymbol{\hat{\sigma}}} \in \R^L$ for better estimation of the relevant axes. The weight vector $\w{}_{\boldsymbol{\hat{\sigma}}}$ is computed as follows:
\begin{align}\label{eq:ARD-VAE_rel_axes_estimation}
    &\w{}_{\boldsymbol{\hat{\sigma}}} = \sum_{k=1}^{D}\frac{1}{N}\sum_{i=1}^{N}\|\mathbb{J}_{i}\|^2, \\ \nonumber
    \text{ where } \mathbb{J} &= \left[ \dfrac{\partial \hat{\x{}}}{\partial \mu_{1}} \cdots \dfrac{\partial \hat{\x{}}}{\partial \mu_{L}} \right] \in \R^{D \times L} \text{ is the Jacobian matrix}.
\end{align}
To get a reliable estimate of the weight vector $\w{}_{\boldsymbol{\hat{\sigma}}}$ using \cref{eq:ARD-VAE_rel_axes_estimation}, we compute the Jacobian for multiple data samples, which is typically the size of a minibatch ($100$) in this work. The weighted estimated variance defined as
\begin{align}\label{eq:ARD-VAE_wt_rel_axes_estimation}
&\boldsymbol{\hat{\sigma}}^2_{\w{}} = \w{}_{\boldsymbol{\hat{\sigma}}} \odot \boldsymbol{\hat{\sigma}}^2
\end{align}
gives us the \emph{relevance score} that is used to determine the relevant axes of the ARD-VAE. The relevance score squashes the variance for the spurious latent axes (removes noise) and relatively scales the variance of the relevant dimensions, as shown in \cref{fig:MNIST_analysis}(ii), and the scores for different $L$s are almost similar. Thus, $\boldsymbol{\hat{\sigma}}^2_{\w{}}$ is better suited to find the relevant/necessary dimensions using the percentage-based approach, typically explaining $99\%$ of the variance in $\boldsymbol{\hat{\sigma}}^2_{\w{}}$ for this work. The importance of latent axes estimated using \cref{eq:ARD-VAE_rel_axes_estimation} can also be used to determine the relevant axes of a trained VAE and its variants (refer to the supp).

\begin{table*}[h]
    \centering
    \resizebox{\textwidth}{!}
    {%
    \begin{tabular}{c c c c c c c c c}
    \hline
    \multirow{2}{*}{\textsc{Method}}&&\multicolumn{3}{c}{\textsc{DSprites}}&&\multicolumn{3}{c}{\textsc{3D Shapes}}\\ 
    \cline{3-6}\cline{7-9}
    &&\textsc{FactorVAE metric}\;\;$\uparrow$&\textsc{MIG}\;\;$\uparrow$ &\textsc{Active}&&\textsc{FactorVAE metric}\;\;$\uparrow$&\textsc{MIG}\;\;$\uparrow$&\textsc{Active}\\ \hline
    VAE $(L=6)$ 
    && $\underline{64.78 \pm 8.05}$ & $0.06 \pm 0.02$ & $6.00 \pm 0.00$
    && $55.85 \pm 8.89$ & $0.13 \pm 0.15$ & $6.00 \pm 0.00$\\
    $\beta$-TCVAE $(L=6)$ 
    && $\mathbf{75.55 \pm 3.52}$ & $\underline{0.20 \pm 0.06}$ & $6.00 \pm 0.00$
    && $\underline{75.51 \pm 12.84}$ & $\underline{0.40 \pm 0.22}$ & $6.00 \pm 0.00$\\
    DIP-VAE-I $(L=6)$ 
    && $59.47 \pm 2.76$ & $0.05 \pm 0.03$ & $6.00 \pm 0.00$
    && $51.94 \pm 1.91$ & $0.06 \pm 0.02$ & $6.00 \pm 0.00$\\
    DIP-VAE-II $(L=6)$ 
    && $60.70 \pm 10.97$ & $0.08 \pm 0.04$ & $6.00 \pm 0.00$
    && $63.66 \pm 11.26$ & $0.24 \pm 0.18$ & $6.00 \pm 0.00$\\
    RAE $(L=6)$ 
    && $64.21 \pm 6.69$ & ${0.04 \pm 0.01}$ & $6.00 \pm 0.00$
    && $53.57 \pm 13.14$ & $0.03 \pm 0.02$ & $6.00 \pm 0.00$\\
    ARD-VAE $(L=10)$ && $63.37 \pm 11.81$ & $\mathbf{0.22 \pm 0.09}$ & $5.80 \pm 0.40$
    && $\mathbf{79.26 \pm 9.31}$ & $\mathbf{0.52 \pm 0.25}$ & $6.40 \pm 0.49$\\  
    \hline
    \end{tabular}
    }
    \caption{Disentanglement scores of competing methods trained with $5$ different seeds on multiple datasets (higher is better). \textsc{Active} indicates the number of the latent dimensions used by different methods. The \textbf{best} score is in \textbf{bold} and the \underline{second best} score is \underline{{underlined}}.} \label{tab:Disentanglement_study}
\end{table*}

\section{Experiments}

\textbf{Datasets : } The DSprites \cite{dsprites17} and 3D Shapes \cite{3dshapes18} datasets are synthetic datasets with \emph{known} latent factors of generation, which are particularly curated for the disentanglement analysis \cite{Dis_benchmark_ICML_2019, TC_VAE_Neurips_2019, Factor_VAE_ICML_2018}. The known latent factors in these datasets serve as the \emph{ground truth} for the number of relevant axes estimated by the ARD-VAE on these datasets. In addition, we study the efficacy of the proposed method on multiple \emph{real-world}
datasets, such as MNIST \cite{MNIST}, CelebA \cite{CelebA}, CIFAR10 \cite{CIFAR10}, and ImageNet ($32\times32$ resolution) \cite{ImageNet} with \emph{unknown} number of latent generative factors.

\textbf{Competing Methods : } To demonstrate the efficacy of the ARD-VAE over other SOTA DLVMs, we use the baseline VAE \cite{kingma2014auto}, $\beta$-TCVAE \cite{TC_VAE_Neurips_2019}, RAE \cite{RAE}, WAE \cite{tolstikhin2017wasserstein}, and DIP-VAE \cite{DIP_VAE_ICLR_2018} in our study for disentanglement analysis on synthetic datasets and modeling the data distributions of real-world datasets. We consider the GECO-$L_0$-ARM-VAE \cite{vae_prune_L0_reg_2021} and MaskAAE \cite{vae_prune_mask_AAE_reg_2020} in our analysis of real-world datasets \emph{only} as the models are highly sensitive to the choice of neural network and hyperparameters.

\textbf{Evaluation Metrics : } In the disentanglement analysis, we evaluate whether the latent dimensions of a DLVM represent true generative factors of the observed data. In an ideal scenario, each latent axis must correspond to a single generative factor. We consider two popular metrics, (i) the FactorVAE metric \cite{Factor_VAE_ICML_2018} and (ii) the mutual information gap (MIG) \cite{TC_VAE_Neurips_2019} in our analysis. Comparison of the model's data distribution and the true but unknown data distribution is challenging, and several methods have been proposed to compare distributions \cite{salimans2016improved,IS_FID_1,precision_recall_distributions}. We use the Fréchet Inception Distance (FID) \cite{IS_FID_1} and precision-recall scores \cite{precision_recall_distributions} to quantify the quality of the samples produced by a model.

\textbf{Implementation Details : } Different neural network architectures and model-specific hyperparameters are used for training a DLVM, depending on the complexity of the data \cite{RAE, Dis_benchmark_ICML_2019, GECO_2018, vae_prune_mask_AAE_reg_2020}. For a fair comparison of the competing methods, we fix the number of latent dimensions, neural network architecture, and optimization settings for a dataset. Details of the neural network architectures and related hyperparameter settings used for different benchmark datasets are reported in the supp. All models are trained with $5$ different initialization seeds for every dataset studied in this work.

\subsection{Results}\label{sec:results}
\subsubsection{Synthetic datasets with \emph{known} latent factors}\label{sec:synthetic_results}
We know the number of latent factors used in the generation of the synthetic datasets, the DSprites \cite{dsprites17} and 3D Shapes \cite{3dshapes18}, which is $6$ for both datasets. Thus, the number of \emph{known} latent factors in these datasets serves as the \emph{ground truth} for the number of relevant (or \textsc{active}) dimensions estimated by the ARD-VAE. In this experiment, we set the initial size of the latent space, $L=10$, and train the ARD-VAE to estimate the number of relevant axes for both datasets. From the results reported in \cref{tab:Disentanglement_study}, we observe the number of active dimensions estimated by the ARD-VAE closely matches the ground truth (GT). Additional results in the supp with $L=\{15,20,30\}$ also match the GT.

\begin{table*}
    \centering
    \resizebox{\textwidth}{!}
    {
    \begin{tabular}{c c c c c c c c c c c}
    \hline
    &&\multicolumn{4}{c}{MNIST\;\;$(L=16)$}&&\multicolumn{4}{c}{CIFAR10\;\;$(L=128)$}\\ \cline{3-6}\cline{8-11}
 &&\textsc{Active} &\textsc{FID}$\downarrow$ &\textsc{Precision}$\uparrow$ &\textsc{Recall}$\uparrow$
 &&\textsc{Active} &\textsc{FID}$\downarrow$ &\textsc{Precision}$\uparrow$ &\textsc{Recall}$\uparrow$\\ \hline
    VAE \
    && $16$ &$28.78 \pm 0.48$ & ${0.88 \pm 0.04}$ & $\underline{0.97 \pm 0.00}$
    && $128$ &$147.74 \pm 0.81$ &$0.50 \pm 0.03$ &$\underline{0.47 \pm 0.02}$ \\
    $\beta$-TCVAE 
    &&  $16$  &$50.62 \pm 1.19$ & $0.82 \pm 0.03$ &$0.95 \pm 0.01$
    &&  $128$  &$180.94 \pm 1.16$ & $0.30 \pm 0.01$ &$0.41 \pm 0.03$ \\
    RAE 
    &&  $16$  & $\mathbf{18.79 \pm 0.31}$ &$0.87 \pm 0.01$ &$0.95 \pm 0.02$ 
    &&  $128$  & $\underline{94.34 \pm 1.58}$ &$\underline{0.74 \pm 0.02}$ &$\underline{0.47 \pm 0.04}$ \\
    WAE 
    &&  $16$  &$25.42 \pm 1.19$ &$\mathbf{0.92 \pm 0.03}$ &$0.92 \pm 0.01$
    &&  $128$  &$140.49 \pm 0.64$ &$0.42 \pm 0.01$ &$0.31 \pm 0.05$ \\
    GECO-$L_0$-ARM-VAE 
    &&$10.00 \pm 1.10$ &$304.75 \pm 64.29$ &$0.03 \pm 0.02$  &$0.38 \pm 0.08$
    &&$68.00 \pm 2.97$ &$320.75 \pm 45.09$ &$0.02 \pm 0.03$ &$0.04 \pm 0.06$\\
    MaskAAE
    &&$9.80 \pm 1.60$ & $144.92 \pm 16.80$ & $0.00 \pm 0.00$ & $0.07 \pm 0.00$
    &&$3.80 \pm 0.40$ & $298.30 \pm 8.44$ & $0.07 \pm 0.02$ & $0.04 \pm 0.05$ \\
    ARD-VAE 
    && $12.80 \pm 0.40$ & $\underline{22.24 \pm 0.57}$ & $\underline{0.91 \pm 0.02}$ & $\mathbf{0.98 \pm 0.00}$
    && $105.80 \pm 1.33$ & $\mathbf{87.56 \pm 1.21}$ & $\mathbf{0.82 \pm 0.03}$ & $\mathbf{0.51 \pm 0.02}$ \\
    [1.1ex] \hline
    \end{tabular}
    }
    \caption{The FID and precision-recall scores of the generated samples along with the number of latent dimensions (\textsc{Active}) used by the competing methods to compute different metric scores. The \textbf{best} score under a metric is in \textbf{bold} and the \underline{second best} score is \underline{{underlined}}.} \label{tab:FID_Score_complete}
\end{table*}

\textbf{Interpretation of the Relevant Axes Discovered by the ARD-VAE: }\label{heading:disentanglement}
It is desirable that the relevant axes identified by VAEs encode meaningful information and represent, to some degree, relevant generative factors for a particular dataset. To evaluate this property, we refer to disentanglement analysis \cite{Dis_benchmark_ICML_2019, TC_VAE_Neurips_2019, Factor_VAE_ICML_2018}.
In addition to the baseline VAE and RAE (which produces good FID scores in \cref{tab:FID_Score_complete}), we consider SOTA VAEs \cite{DIP_VAE_ICLR_2018,TC_VAE_Neurips_2019}, which are explicitly designed to produce disentangled latent representations in our analysis. For all the competing methods, except the ARD-VAE, the size of the latent space ($L=6$) is set to the number of the {\em known latent factors} of the DSprites \cite{dsprites17} and Shapes3D \cite{3dshapes18} dataset --- a significant advantage, which would not be available in most applications. We follow the encoder-decoder architectures and training strategies used in \cite{Dis_benchmark_ICML_2019} for all the methods.

From the analysis reported in \cref{tab:Disentanglement_study}, we observe that the ARD-VAE produces the best disentangled representation for the Shapes3D dataset when evaluated using both metrics. The precise identification of the true generative factors by the ARD-VAE for this dataset helps achieve such scores. The best MIG score of the ARD-VAE on the DSprites dataset indicates that the representations learned by the relevant axes are not entangled relative to other methods. We know that the FactorVAE metric score does not penalize a method if a ground truth factor is represented by multiple latent axes, unlike the MIG. Thus, we observe a low MIG score in many scenarios for methods with a relatively high score for the FactorVAE metric, such as the VAE, DIP-VAE, and RAE. Though the RAE does a reasonable job of learning the data distribution (in \cref{tab:FID_Score_complete}), the learned latent representations are not disentangled, as indicated by low scores. 
The performance of the DIP-VAE is often poorer than the baseline VAE despite being designed to learn disentangled representations. Among the different variants of the VAE, the $\beta$-TCVAE produces results comparable to the ARD-VAE.

\subsubsection{Real datasets with \emph{unknown} latent factors}\label{sec:realdataset_results}
For the real datasets, we do not know the number of latent factors and is determined empirically using cross-validation for an encoder-decoder model. As the \emph{true/intrinsic} dimension of a real dataset is unknown, we compare the data distribution modeled by a method with the true but unknown data distribution (using samples from both distributions) as a proxy to evaluate the effectiveness of a method (for a given bottleneck size). In this experiment, we show that the relevant axes determined by the ARD-VAE produce better or comparable performance under different evaluation metrics. Moreover, the proposed method is not sensitive to the choice of neural networks or other hyperparameters.

\textbf{Evaluation of the Learned Data Distribution: } The initial latent dimensions used for the MNIST, CelebA, CIFAR10, and ImageNet datasets are $L=16$, $L=64$, $L=128$, and $L=256$, respectively \cite{tolstikhin2017wasserstein,RAE}. The motivation of this analysis is to examine whether the proposed ARD-VAE is able to find the relevant axes required for modeling a dataset using the standard SOTA encoder-decoder architectures and to examine how close the learned data distribution is to the true distribution. The estimated variance in \cref{eq:ARD-VAE_estimated_var} is used to generate samples for the ARD-VAE. Besides, the FID scores of the generated samples, reported as \textsc{FID}, we report the precision-recall scores under \textsc{precision} and \textsc{recall}, respectively, in \cref{tab:FID_Score_complete}. Results for the CelebA and ImageNet datasets are reported in the supp. The number of latent dimensions used by a method to reconstruct/generate samples are reported as \textsc{Active}. Using the importance of the latent axes estimated using \cref{eq:ARD-VAE_rel_axes_estimation}, we determine the relevant axes even for the VAE, $\beta$-TCVAE, RAE, and WAE that are used for model evaluation (results in the supp).

The relevant dimensions estimated by the ARD-VAE in \cref{tab:FID_Score_complete} are consistently less than the initial (or nominal) dimension $L$ for all the datasets. We have no knowledge of any "ground truth" intrinsic dimension for these datasets. Instead, we rely on comparisons with empirically derived bottle-neck sizes from literature and evaluate whether the subset of learned dimensions is sufficient to represent or reproduce the data. These results show that the performance of the ARD-VAE is better than the baseline VAE. The WAE-MMD ignores the higher order moments in matching distributions \cite{MMD_2012}, possibly leading to \emph{holes} in the encoded distribution and resulting in higher FID and lower precision-recall scores. The RAE performs reasonably well across multiple datasets under different scenarios, conceivably due to the use of aggregate latent distribution to generate samples \cite{tolstikhin2017wasserstein,dai2019_2sVAE}. However, the learned representations cannot be readily used for subsequent statistical analysis, such as the disentanglement analysis (poor scores in \cref{tab:Disentanglement_study}) and outlier detection \cite{gens_saha_2022}, due to the lack of structure in the latent representations.

The MaskAAE and GECO-$L_0$-ARM-VAE, which are designed to find the relevant axes to represent a dataset using DLVMs, have failed, using this encoder-decoder architecture, to learn the data distribution, as indicated by significantly high \textsc{FID} and low precision-recall scores relative to other methods. This result may signal issues about the generalization of those formulations and sensitivity to the hyperparameter tuning relative to different architectures and datasets (as will be further examined in the following analysis). The number of relevant (or \textsc{active}) dimensions estimated by the GECO-$L_0$-ARM-VAE are sufficient to produce reasonably good reconstructed samples, indicated by comparable reconstruction loss (reported in the supp). However, using this encoder-decoder architecture, it failed to map samples to the prior distribution, $\Gauss{\zero}{\Id}$, in the latent space. In contrast, the number of relevant dimensions estimated by the MaskAAE collapsed for the complex datasets, such as the CelebA (reported in the supp) and CIFAR10. We have tuned the hyperparameters of the MaskAAE over a series (i.e., dozens) of experiments, and no better results could be obtained. The number of hyperparameters corresponding to different regularization losses added to the AAE loss, associated with a complex training recipe, makes the tuning of the MaskAAE difficult. Compared to the MaskAAE and GECO-$L_0$-ARM-VAE, the ARD-VAE could find the relevant dimensions required to model the data distribution, supported by favorable (i.e., the best or nearly best) metric scores. This experiment demonstrates the ability of the ARD-VAE to find relevant axes and learn the distribution of real-world datasets.

\begin{table}
    \centering
    \resizebox{0.49\textwidth}{!}
    {
    \begin{tabular}{c c c c c c c}
    \hline
    &&\multicolumn{2}{c}{$2L $}&&\multicolumn{2}{c}{$4L $}\\ 
    \cline{3-4}\cline{6-7}
    &&\textsc{Active}&\textsc{FID} $\downarrow$ &&\textsc{Active}&\textsc{FID}  $\downarrow$\\ \hline
    MNIST $(L=16)$ 
    && $12.60 \pm 0.49$ & $22.30 \pm 0.33$
    && $12.40 \pm 0.49$ & $22.31 \pm 0.58$ \\
    CIFAR10 $(L=128)$ 
    && $116.40 \pm 3.72$ & $86.50 \pm 1.43$ 
    && $117.80 \pm 15.04$ & $87.88 \pm 1.8$ \\
    \hline
    \end{tabular}
    }
    \caption{The number of active dimensions (\textsc{Active}) discovered by the ARD-VAE and the FID scores of the generated samples (\textsc{FID}). The bottleneck dimension is specified as a multiples of $L$.} 
    \label{tab:FID_Score_Latent_Size}
\end{table}

\textbf{Sensitivity of $\beta$ to the Initial Bottleneck Size $L$ in the ARD-VAE: } In real applications, we would not know the intrinsic dimension of a dataset. Thus, the interesting use case of this methodology is to start with a relatively high-dimensional latent space and let the algorithm/training find the relevant dimensions required to model a particular dataset. For a given latent space size, the hyperparameter $\beta$ of the ARD-VAE is regulated to achieve an approximate reconstruction loss. Under this scenario, it would be interesting to examine whether the value of $\beta$ is dependent on the latent space's initial size. To understand this phenomenon, we devised an experiment where we fix $\beta$ (for a dataset and an autoencoder architecture) and vary the size of the latent space. We evaluate the ARD-VAE for the number of estimated relevant dimensions and the FID scores of the learned data distribution for the MNIST and CIFAR10 datasets. The results in \cref{tab:FID_Score_Latent_Size} indicate that the hyperparameter $\beta$ of the ARD-VAE is agnostic to the initial size of the latent space, in general. The outcome is consistent with the results on synthetic datasets, reported in the supp.


\begin{table}
    \resizebox{0.49\textwidth}{!}
    {
    \begin{threeparttable}
    \begin{tabular}{c c c c c c}
    \hline
    &\multicolumn{2}{c}{MaskAAE}&&\multicolumn{2}{c}{ARD-VAE}\\ \cline{2-3}\cline{5-6}
    &\textsc{Active}&\textsc{FID} $\downarrow$&&\textsc{Active}&\textsc{FID} $\downarrow$\\ \hline
    MNIST$(L=16)$ & $9.80 \pm 2.23$ & $\mathbf{21.35 \pm 0.70}$
        &&  $8.0 \pm 0.0$ & $21.74 \pm 0.49$ \\
    MNIST$(l=32)$ & $11.40 \pm 2.65$ & $21.74 \pm 3.77$ &&  $7.8 \pm 0.4$ & $\mathbf{20.32 \pm 0.96}$ \\
    CIFAR$(L=256)$\tnote{1} & NA & NA &&  $60.4 \pm 10.89$ & $94.86 \pm 2.68$ \\ \hline
    \end{tabular}
    \begin{tablenotes}
    \item[1] The MaskAAE collapsed all the dimensions for the CIFAR10 dataset after $20$ epochs. Thus, we skip the evaluation of the MaskAAE for the CIFAR10 dataset.
    \end{tablenotes}
    \end{threeparttable}
    }
    \caption{The FID scores of the generated samples (\textsc{FID}) using the active dimensions (\textsc{Active}) identified by the MaskAAE and ARD-VAE when trained using the neural network architectures used in the MaskAAE \cite{vae_prune_mask_AAE_reg_2020}.} \label{tab:FID_Score_Mask_AAE_Architecture}
\end{table}

\begin{table}
    \resizebox{0.49\textwidth}{!}
    {
    \begin{tabular}{c c c c c c}
    \hline
    &\multicolumn{2}{c}{GECO-$L_0$-ARM-VAE}&&\multicolumn{2}{c}{ARD-VAE}\\ \cline{2-3}\cline{5-6}
&\textsc{Active}&$\textsc{FID} \downarrow$&&\textsc{Active}&$\textsc{FID} \downarrow$\\ \hline
    MNIST$(L=16)$ 
    &$9.00 \pm 1.41$ & $64.05 \pm 5.31$
    &&  $14.60 \pm 0.49$ & $\mathbf{34.45 \pm 0.82}$ \\
    MNIST$(L=32)$ 
    &$18.2 \pm 2.93$ & $93.62 \pm 22.49$
    &&  $15.60 \pm 0.49$ & $\mathbf{34.15 \pm 1.15}$ \\
    \hline
    \end{tabular}
    }
    \caption{The number of the active dimensions (\textsc{Active}) and FID scores of the generated samples (\textsc{FID}) of the GECO-$L_0$-ARM-VAE and ARD-VAE using the neural network architectures used in the GECO-$L_0$-ARM-VAE \cite{vae_prune_L0_reg_2021}.} \label{tab:FID_Score_GECO}
\end{table}

\textbf{Performance of the ARD-VAE across Neural Network Architectures: } For the encoder-decoder architecture used in \cref{tab:FID_Score_complete}, we have observed that the MaskAAE and GECO-$L_0$-ARM-VAE could not produce comparable results. This is possibly due to the sensitivity to the neural network architecture and related hyperparameter settings. To understand the dependency of these methods on the architectures and optimization settings used by the authors, we ran these methods to reproduce the results reported in the papers associated with those methods. In table \cref{tab:FID_Score_Mask_AAE_Architecture}, we report the performance of the MaskAAE and ARD-VAE using the network architectures proposed by the authors in \cite{vae_prune_mask_AAE_reg_2020}. In table \cref{tab:FID_Score_GECO}, the encoder-decoder architecture reported in \cite{vae_prune_L0_reg_2021} is used to train the GECO-$L_0$-ARM-VAE and ARD-VAE. 

In \cref{tab:FID_Score_Mask_AAE_Architecture}, the FID scores of the samples generated by the MaskAAE for the MNIST dataset \footnote{The FID scores found in this study, using an online version of the authors' code, are higher than the those reported in \cite{vae_prune_mask_AAE_reg_2020}.} are close to the scores of other methods in \cref{tab:FID_Score_complete}. These results demonstrate the efficacy of MaskAAE with the proper choice of neural network architecture, hyperparameters, and optimization strategy. Despite that, we could not train the MaskAAE on the CIFAR10 dataset using the architecture and hyper-parameter settings reported by those authors; for the CIFAR10, the MaskAAE collapsed all the latent dimensions. In contrast, we could successfully train the ARD-VAE on both the MNIST and CIFAR10 by finding a reasonable value for the $\beta$ (without excessive fine-tuning) that produces good reconstruction.

From the results reported in \cref{tab:FID_Score_GECO}, we observe that the relevant dimensions  estimated by the GECO-$L_0$-ARM-VAE are sensitive to the size of the initial latent dimension, $L$, even with a fixed $\tau$ (target reconstruction loss used in \cite{vae_prune_L0_reg_2021}). However, the estimated ARD-VAE for a fixed $\beta$ is indifferent to $L$ and estimates almost the same number of relevant dimensions. Moreover, the ARD-VAE does better in learning the data distribution (i.e., lower FID scores). Relatively higher FID scores using this architecture for the MNIST dataset are likely due to the use of non-trainable upsampling layers in the decoder.

\section{Conclusion}
In this work, we present a statistical approach to determine the relevant axes in the VAE using a hierarchical prior without deviating from and adding any regularization loss to the ELBO formulation. Unlike other relevancy detection methods for DLVMs \cite{vae_prune_mask_AAE_reg_2020,vae_prune_L0_reg_2021}, the ARD-VAE is resilient to the choice of neural network architectures and is impervious to the size of the latent space.
Extensive empirical evaluations supported by comprehensive ablation studies demonstrate the 
\emph{stability, robustness}, and \emph{effectiveness} of the proposed ARD-VAE. This work is supported by the National Institutes of Health, grant 5R01ES032810.


{\small
\bibliographystyle{ieee_fullname}
\bibliography{paper}

\begin{thebibliography}{10}\itemsep=-1pt

\bibitem{Berger_Statistics_Gamma}
James~O. Berger.
\newblock {\em Statistical decision theory and Bayesian analysis}.
\newblock Springer, 2nd edition, 1985.

\bibitem{Bishop_Bayesian_PCA_1998}
Christopher Bishop.
\newblock Bayesian pca.
\newblock In M. Kearns, S. Solla, and D. Cohn, editors, {\em Advances in Neural Information Processing Systems}, volume~11. MIT Press, 1998.

\bibitem{Bishop_PRML}
Christopher~M. Bishop.
\newblock {\em Pattern Recognition and Machine Learning}.
\newblock Springer, 8th edition, 2009.

\bibitem{3dshapes18}
Chris Burgess and Hyunjik Kim.
\newblock 3d shapes dataset.
\newblock https://github.com/deepmind/3d-shapes/, 2018.

\bibitem{dai2019_2sVAE}
Bin Dai and David Wipf.
\newblock Diagnosing and enhancing vae models.
\newblock {\em International Conference on Learning Representations}, 2019.

\bibitem{vae_prune_L0_reg_2021}
Cedric De~Boom, Samuel Wauthier, Tim Verbelen, and Bart Dhoedt.
\newblock Dynamic narrowing of vae bottlenecks using geco and l0 regularization.
\newblock In {\em International Joint Conference on Neural Networks (IJCNN)}, 2021.

\bibitem{ImageNet}
Jia Deng, Wei Dong, Richard Socher, Li-Jia Li, Kai Li, and Li Fei-Fei.
\newblock Imagenet: A large-scale hierarchical image database.
\newblock In {\em 2009 IEEE Conference on Computer Vision and Pattern Recognition}, pages 248--255, 2009.

\bibitem{RAE}
Partha Ghosh, Mehdi S.~M. Sajjadi, Antonio Vergari, Michael Black, and Bernhard Scholk\"{o}pf.
\newblock From variational to deterministic autoencoders.
\newblock In {\em International Conference on Learning Representations}, 2020.

\bibitem{MMD_2012}
Arthur Gretton, Karsten~M. Borgwardt, Malte~J. Rasch, Bernhard Scholkopf, and Alexander Smola.
\newblock A kernel two-sample test.
\newblock {\em Journal of Machine Learning Research}, 13:723--773, 2012.

\bibitem{vae_basic_syn_mol_2018}
Rafael Gómez-Bombarelli, Jennifer~N. Wei, David Duvenaud, José~Miguel Hernández-Lobato, Benjamín Sánchez-Lengeling, Dennis Sheberla, Jorge Aguilera-Iparraguirre, Timothy~D. Hirzel, Ryan~P. Adams, and Alán Aspuru-Guzik.
\newblock Automatic chemical design using a data-driven continuous representation of molecules.
\newblock {\em ACS Central Science}, 4(2):268--276, 2018.

\bibitem{vae_few_shot_2023}
Jiaming Han, Yuqiang Ren, Jian Ding, Ke Yan3, and Gui-Song Xia.
\newblock Few-shot object detection via variational feature aggregation.
\newblock In {\em AAAI Conference on Artificial Intelligence}, 2023.

\bibitem{IS_FID_1}
Martin Heusel, Hubert Ramsauer, Thomas Unterthiner, Bernhard Nessler, and Sepp Hochreiter.
\newblock Gans trained by a two time-scale update rule converge to a local nash equilibrium.
\newblock In {\em Conference on Neural Information Processing Systems}, 2017.

\bibitem{beta_VAE_ICLR_2017}
Irina Higgins, Loic Matthey, Arka Pal, Christopher Burgess, Xavier Glorot, Matthew Botvinick, Shakir Mohamed, and Alexander Lerchner.
\newblock $\beta$-vae: Learning basic visual concepts with a constrained variational framework.
\newblock In {\em International Conference on Learning Representations}, 2017.

\bibitem{image_gen_diffusion}
Jonathan Ho, Ajay Jain, and Pieter Abbeel.
\newblock Denoising diffusion probabilistic models.
\newblock In {\em Advances in Neural Information Processing Systems (NeurIPS)}, 2020.

\bibitem{ELBO_surgery_2016}
Matthew~D. Hoffman and Matthew~J. Johnson.
\newblock Elbo surgery: yet another way to carve up the variational evidence lower bound.
\newblock In {\em NIPS Workshop: Advances in Approximate Bayesian Inference}, 2016.

\bibitem{kachigan1982multivariate}
S.K. Kachigan.
\newblock {\em Multivariate Statistical Analysis: A Conceptual Introduction}.
\newblock Radius Press, 1982.

\bibitem{Style_GAN_2021}
Tero Karras, Miika Aittala, Samuli Laine, Erik Härkönen, Janne Hellsten, Jaakko Lehtinen, and Timo Aila.
\newblock Alias-free generative adversarial networks.
\newblock In {\em Conference on Neural Information Processing Systems}, 2021.

\bibitem{Factor_VAE_ICML_2018}
Hyunjik Kim and Andriy Mnih.
\newblock Disentangling by factorising.
\newblock In {\em International Conference on Machine Learning}, 2018.

\bibitem{kingma2014auto}
Diederik~P Kingma and Max Welling.
\newblock Auto-encoding variational bayes.
\newblock {\em International Conference on Learning Representations}, 2014.

\bibitem{CIFAR10}
Alex Krizhevsky.
\newblock Learning multiple layers of features from tiny images, 2009.
\newblock Available on: \url{https://www.cs.toronto.edu/~kriz/cifar.html}.

\bibitem{DIP_VAE_ICLR_2018}
Abhishek Kumar, Prasanna Sattigeri, and Avinash Balakrishnan.
\newblock Variational inference of disentangled latent concepts from unlabeled observations.
\newblock In {\em International Conference on Learning Representations}, 2018.

\bibitem{MNIST}
Yann LeCun, Corinna Cortes, and CJ Burges.
\newblock Mnist handwritten digit database, 2010.
\newblock Available on: \url{http://yann.lecun.com/exdb/mnist}.

\bibitem{L0_ARM_L0_reg_2019}
Yang Li and Shihao Ji.
\newblock L0-arm: Network sparsification via stochastic binary optimization.
\newblock In {\em European Conference on Machine Learning}, 2019.

\bibitem{vae_semi_sup_2019}
Yang Li, Quan Pan, Suhang Wang, Haiyun Peng, Tao Yang, and Erik Cambria.
\newblock Disentangled variational auto-encoder for semi-supervised learning.
\newblock {\em Information Sciences}, 482:73--85, 2019.

\bibitem{CelebA}
Ziwei Liu, Ping Luo, Xiaogang Wang, and Xiaoou Tang.
\newblock Deep learning face attributes in the wild.
\newblock In {\em International Conference on Computer Vision}, 2015.

\bibitem{Dis_benchmark_ICML_2019}
Francesco Locatello, Stefan Bauer, Mario Lucic, Gunnar Rätsch, Sylvain Gelly, Bernhard Schölkopf, and Olivier Bachem.
\newblock Challenging common assumptions in the unsupervised learning of disentangled representations.
\newblock In {\em International Conference on Machine Learning}, 2019.

\bibitem{posterior_collapse_Neurips_2019}
James Lucasz, George Tuckery, Roger Grossez, and Mohammad Norouziy.
\newblock Don’t blame the elbo! a linear vae perspective on posterior collapse.
\newblock In {\em Conference on Neural Information Processing Systems}, 2019.

\bibitem{posterior_collapse_ICLR_2019}
James Lucasz, George Tuckery, Roger Grossez, and Mohammad Norouziy.
\newblock Understanding posterior collapse in generative latent variable models.
\newblock In {\em International Conference on Learning Representations}, 2019.

\bibitem{Bishop_Bayesian_PCA_1999}
Christopher M.~Bishop.
\newblock Variational principal components.
\newblock In {\em ICANN}, 1999.

\bibitem{vae_few_shot_2020}
Peirong Ma and Xiao Hu.
\newblock A variational autoencoder with deep embedding model for generalized zero-shot learning.
\newblock In {\em AAAI Conference on Artificial Intelligence}, 2020.

\bibitem{makhzani2016adversarial}
Alireza Makhzani, Jonathon Shlens, Navdeep Jaitly, Ian Goodfellow, and Brendan Frey.
\newblock Adversarial autoencoders.
\newblock In {\em International Conference on Learning Representations}, 2016.

\bibitem{dsprites17}
Loic Matthey, Irina Higgins, Demis Hassabis, and Alexander Lerchner.
\newblock dsprites: Disentanglement testing sprites dataset.
\newblock https://github.com/deepmind/dsprites-dataset/, 2017.

\bibitem{vae_prune_mask_AAE_reg_2020}
Arnab~Kumar Mondal, Sankalan~Pal Chowdhury, Aravind Jayendran, Parag Singla, Himanshu Asnani, and Prathosh AP.
\newblock Maskaae: Latent space optimization for adversarial auto-encoders.
\newblock In {\em Uncertainty in Artificial Intelligence (UAI)}, 2020.

\bibitem{muirhead1982aspects}
R.J. Muirhead.
\newblock {\em Aspects of Multivariate Statistical Theory}.
\newblock Wiley Series in Probability and Statistics. Wiley, 1982.

\bibitem{Baeyesian_conjugate}
Kevin~P. Murphy.
\newblock {\em Conjugate Bayesian analysis of the Gaussian distribution}.
\newblock Technical report, 2007.

\bibitem{Neal_ARD_1996}
Radford~M. Neal.
\newblock {\em Bayesian Learning for Neural Networks}.
\newblock Springer, 1996.

\bibitem{vae_basic_mol_struct_2023}
Toshiki Ochiai, Tensei Inukai, Manato Akiyama, Kairi Furui, Masahito Ohue, Nobuaki Matsumori, Shinsuke Inuki, Motonari Uesugi, Toshiaki Sunazuka, Kazuya Kikuchi, Hideaki Kakeya, and Yasubumi Sakakibara.
\newblock Variational autoencoder-based chemical latent space for large molecular structures with 3d complexity.
\newblock {\em Communications Chemistry}, 6(249), 2023.

\bibitem{posterior_collapse_VQ_VAE_2017_NIPS}
Aaron van~den Oord, Oriol Vinyals, and Koray Kavukcuoglu.
\newblock Neural discrete representation learning.
\newblock In {\em Conference on Neural Information Processing Systems}, 2017.

\bibitem{vae_basic_rna_seq_2023}
Alessandro Palma, Sergei Rybakov, Leon Hetzel, and Fabian Theis.
\newblock Modelling single-cell rna-seq trajectories on a flat statistical manifold.
\newblock In {\em NeurIPS AI for Science Workshop}, 2023.

\bibitem{posterior_collapse_delta_VAE_ICLR_2019}
Ali Razavi, Aaron van~den Oord, Ben Poole, and Oriol Vinyals.
\newblock Preventing posterior collapse with $\delta$-vaes.
\newblock In {\em International Conference on Learning Representations}, 2019.

\bibitem{rezende2014stochastic}
Danilo~Jimenez Rezende, Shakir Mohamed, and Daan Wierstra.
\newblock Stochastic backpropagation and approximate inference in deep generative models.
\newblock In {\em International Conference on Machine Learning}, pages 1278--1286, 2014.

\bibitem{GECO_2018}
Danilo~J. Rezende and Fabio Viola.
\newblock Generalized elbo with constrained optimization, geco.
\newblock In {\em Third workshop on Bayesian Deep Learning (NeurIPS)}, 2018.

\bibitem{rosca2018distribution}
Mihaela Rosca, Balaji Lakshminarayanan, and Shakir Mohamed.
\newblock Distribution matching in variational inference.
\newblock {\em arxiv}, 2018.
\newblock Preprint at \url{https://arxiv.org/abs/1802.06847}.

\bibitem{gens_saha_2022}
Surojit Saha, Shireen Elhabian, and Ross Whitaker.
\newblock Gens: generative encoding networks.
\newblock {\em Machine Learning}, 111:4003–4038, 2022.

\bibitem{avae_saha_2023}
Surojit Saha, Sarang Joshi, and Ross Whitaker.
\newblock Matching aggregate posteriors in the variational autoencoder, 2023.
\newblock Preprint at \url{https://arxiv.org/pdf/2311.07693.pdf}.

\bibitem{precision_recall_distributions}
Mehdi~S.~M. Sajjadi, Olivier Bachem, Mario Lu{\v c}i{\'c}, Olivier Bousquet, and Sylvain Gelly.
\newblock {Assessing Generative Models via Precision and Recall}.
\newblock In {\em {Advances in Neural Information Processing Systems (NeurIPS)}}, 2018.

\bibitem{salimans2016improved}
Tim Salimans, Ian Goodfellow, Wojciech Zaremba, Vicki Cheung, Alec Radford, and Xi Chen.
\newblock Improved techniques for training gans.
\newblock In {\em Conference on Neural Information Processing Systems}, 2016.

\bibitem{vae_few_shot_2019}
Edgar Schonfeld, Sayna Ebrahimi, Samarth Sinha, Trevor Darrell, and Zeynep Akata.
\newblock Generalized zero- and few-shot learning via aligned variational autoencoders.
\newblock In {\em IEEE Conference on Computer Vision and Pattern Recognition}, 2019.

\bibitem{score_based_diffusion}
Yang Song, Jascha Sohl-Dickstein, Diederik~P. Kingma, Abhishek Kumar, Stefano Ermon, and Ben Poole.
\newblock Score-based generative modeling through stochastic differential equations.
\newblock In {\em International Conference on Learning Representations}, 2021.

\bibitem{TC_VAE_Neurips_2019}
Ricky T.~Q.~Chen, Xuechen Li, Roger Grosse, and David Duvenaud.
\newblock Isolating sources of disentanglement in vaes.
\newblock In {\em Conference on Neural Information Processing Systems}, 2019.

\bibitem{rvm_tipping_2001}
Michael~E. Tipping.
\newblock Sparse bayesian learning and the relevance vector machine.
\newblock {\em Journal of Machine Learning Research}, 1:211--244, 2001.

\bibitem{tolstikhin2017wasserstein}
Ilya Tolstikhin, Olivier Bousquet, Sylvain Gelly, and Bernhard Schoelk\"{o}pf.
\newblock Wasserstein auto-encoders.
\newblock In {\em International Conference on Learning Representations}, 2018.

\bibitem{vae_weak_sup_2020}
Francesco Tonolini, Nikolaos Aletras, Yunlong Jiao, and Gabriella Kazai.
\newblock Nestedvae: Isolating common factors via weak supervision.
\newblock In {\em IEEE Conference on Computer Vision and Pattern Recognition}, 2020.

\bibitem{vae_weak_sup_2023}
Matthew~J. Vowels, Necati~Cihan Camgoz, and Richard Bowden.
\newblock Robust weak supervision with variational auto-encoders.
\newblock In {\em International Conference on Machine Learning}, 2023.

\end{thebibliography}
}

\end{document}


\title{Supplementary to the ARD-VAE: A Statistical Formulation to Find the Relevant Latent Dimensions of Variational Autoencoders}

\author{Surojit Saha\\
The University of Utah, USA\\
{\tt\small surojit.saha@utah.edu}
\and
Sarang Joshi\\
The University of Utah, USA \\
{\tt\small sarang.joshi@utah.edu}
\and
Ross Whitaker\\
The University of Utah, USA \\
{\tt\small whitaker@cs.utah.edu}
}
\maketitle

\section{Training of the ARD-VAE}\label{app:algorithms}
The \cref{alg:ARD-VAE_Alg} outlines the training of the ARD-VAE. The proposed method uses data in the latent space, $D_{\z{}}$, to update the parameters of the prior distribution, $p(\z{} \mid D{}_{\z{}})$ (that is not fixed, unlike the regular VAE \cite{kingma2014auto, rezende2014stochastic}). The data $D_{\z{}}$ in the latent space is produced using the \cref{alg:sample_hyperpriors} that takes in as input a subset of the training data, $\mathcal{X{}}_{\boldsymbol\alpha}$.

\begin{algorithm}
\caption{: \textbf{ARD-VAE training}}
\begin{algorithmic}
    \STATE {\bfseries Input:} Training samples $\mathcal{X}$, Number of epochs the $D_{\z{}}$ should lag $u_{D_{\z{}}}$, Scaling factor $\beta$
    \STATE {\bfseries Output:} Encoder and decoder parameters, $\phi$ and $\theta$ 
    \STATE Split $\mathcal{X{}}$ into training, $\mathcal{X{}}_{train}$, and validation data, $\mathcal{X{}}_{val}$
    \STATE Choose a subset $\mathcal{X{}}_{\boldsymbol\alpha}$ at random from $\mathcal{X{}}_{train}$
    \STATE Initialize SGD samples $\mathcal{X{}}_{sgd} = \mathcal{X{}}_{train} - \mathcal{X{}}_{\boldsymbol\alpha}$
    \STATE Initialize $\phi$ and $\theta$
    \STATE Initialize epoch index $e \gets 0$
    \FOR {number of epochs}
    \IF {$e \mod u_{D_{\z{}}}$}
    \STATE Choose a new subset of the training data $\mathcal{X{}}_{\boldsymbol\alpha}$
    \STATE Update SGD samples, $\mathcal{X{}}_{sgd} = \mathcal{X{}}_{train} - \mathcal{X{}}_{\boldsymbol\alpha}$
    \STATE Update $D_{\z{}}$ with the samples produced by the \cref{alg:sample_hyperpriors} using the update $\mathcal{X{}}_{\boldsymbol\alpha}$
    \STATE Update $\boldsymbol{b}_{L}$, using the latest $D_{\z{}}$
    \ENDIF
    \FOR {number of minibatch updates}
    \STATE Sample a minibatch from $\mathcal{X{}}_{sgd}$
    \STATE Update $\phi$ and $\theta$ by optimizing the ARD-VAE objective function
    \ENDFOR
    \STATE $e \gets e+1$
    \STATE Shuffle $\mathcal{X{}}_{sgd}$
    \ENDFOR
    \end{algorithmic}
\label{alg:ARD-VAE_Alg}
\end{algorithm}

\begin{algorithm}
   \caption{Produce $D_{\z{}}$ using the training subset $\mathcal{X{}}_{\boldsymbol\alpha}$}
   \label{alg:sample_hyperpriors}
\begin{algorithmic}
   \STATE {\bfseries Input:} $\mathcal{X{}}_{\boldsymbol\alpha}$
   \STATE {\bfseries Output:} $D_{\z{}}$
   \STATE Initialize $D_{\z{}} \leftarrow \phi$.
   \FOR {$\x^{'}{} \in \mathcal{X}_{\boldsymbol\alpha}$}
   \STATE $\boldsymbol\mu_{\x^{'}{}}, \boldsymbol\sigma_{\x^{'}{}}^2 \leftarrow \mathbf{E_{\phi}}(\x^{'}{})$
   \STATE $\z^{'}{} \leftarrow \boldsymbol\mu_{\x^{'}{}} + \boldsymbol\epsilon \odot \boldsymbol\sigma_{\x^{'}{}}$
   \STATE $D_{\z{}} \leftarrow D_{\z{}} \cup \z^{'}{}$
   \ENDFOR
\end{algorithmic}
\end{algorithm}

\textbf{Ablation Study on the Size of $\mathcal{X{}}_{\boldsymbol\alpha}$ : } The size of the $\mathcal{X{}}_{\boldsymbol\alpha}$ in \cref{alg:ARD-VAE_Alg} can be treated as a hyperparameter of the proposed method. Thus, we conduct an ablation study on the size of the $\mathcal{X{}}_{\boldsymbol\alpha}$. In this analysis, we study the effect of the number of samples in $\mathcal{X{}}_{\boldsymbol\alpha}$ on the performance of the ARD-VAE in terms of the number of relevant/active dimensions identified and FID score of the generated samples. Different settings of $\mid\mathcal{X{}}_{\boldsymbol\alpha}\mid$ considered in this study are $\mid\mathcal{X{}}_{\boldsymbol\alpha}\mid=\{2K, 5K, 10K, 20K\}$. The ARD-VAE is trained on the MNIST and CIFAR10 datasets with $L=32$ and $L=256$, respectively. We have used the neural network architecture in \cref{tab:Encoder_decoder_NN_1} and followed the optimization strategies discussed in the section \ref{app:benchmark_arch}. We use the hyperparameter $\beta$ as reported in \cref{tab:hyper-parameters}. Compared to other results reported in the results section, the network parameters are initialized using a single seed, as we did not observe much variation in the performance of the ARD-VAE with different initializations.
 
From the results reported in \cref{tab:FID_Score_Gaussian_Samples}, we observe negligible variation in the number of relevant dimensions estimated by the ARD-VAE with the size of $\mathcal{X{}}_{\boldsymbol\alpha}$ and there is almost no variation in the FID scores of the generated samples for both datasets. Thus, we conclude the performance of the ARD-VAE is not affected by the number of samples in $\mathcal{X{}}_{\boldsymbol\alpha}$. The ARD-VAE is trained in an unsupervised framework having access to a large amount of data, and it is good to have a representative set that captures the variations in the dataset. Thus, we choose $\mid\mathcal{X{}}_{\boldsymbol\alpha}\mid=10K$ as a general setting for any random dataset that uses the ARD-VAE. 

\begin{table*}
    \centering
    {%
    \begin{tabular}{c c c c c c c c c c c c c}
    \hline
    &&\multicolumn{2}{c}{$\mid\mathcal{X{}}_{\boldsymbol\alpha}\mid=2K$}&&\multicolumn{2}{c}{$\mid\mathcal{X{}}_{\boldsymbol\alpha}\mid=5K$}&&\multicolumn{2}{c}{$\mid\mathcal{X{}}_{\boldsymbol\alpha}\mid=10K$}&&\multicolumn{2}{c}{$\mid\mathcal{X{}}_{\boldsymbol\alpha}\mid=20K$}\\ \cline{3-4}\cline{6-7}\cline{9-10}\cline{12-13}
    &&\textsc{Active}&\textsc{FID} $\downarrow$ &&\textsc{Active}&\textsc{FID}  $\downarrow$&&\textsc{Active}&\textsc{FID}  $\downarrow$ &&\textsc{Active}&\textsc{FID}  $\downarrow$\\ \hline
    MNIST $(L=32)$
    && $13$ & $21.83$
    && $12$ & $22.04$
    && $13$ & $22.13$ 
    && $13$ & $24.12$ \\
    CIFAR10 $(L=256)$ 
    && $116$ & $85.15$
    && $114$ & $85.65$
    && $112$ & $85.99$ 
    && $112$ & $86.83$ \\
    \hline
    \end{tabular}}
    \caption{The number of \textsc{Active} dimensions and the FID scores of the generated samples for different sizes of the $\mathcal{X{}}_{\boldsymbol\alpha}$ in the ARD-VAE.} \label{tab:FID_Score_Gaussian_Samples}
\end{table*}

\begin{figure*}
  \begin{subfigure}{0.32\linewidth}
    \centerline{\includegraphics[width=0.95\textwidth]{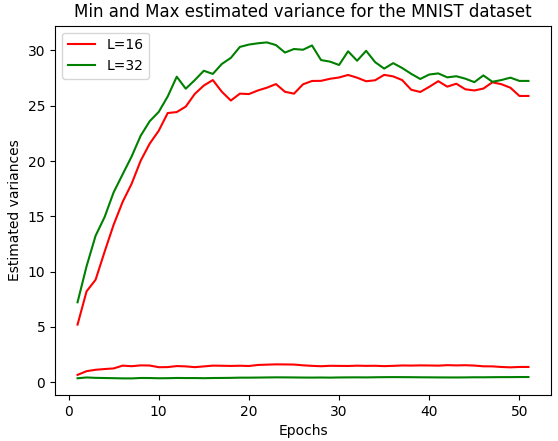}}
    \caption{MNIST}
  \end{subfigure}
  \begin{subfigure}{0.32\linewidth}
    \centerline{\includegraphics[width=0.95\textwidth]{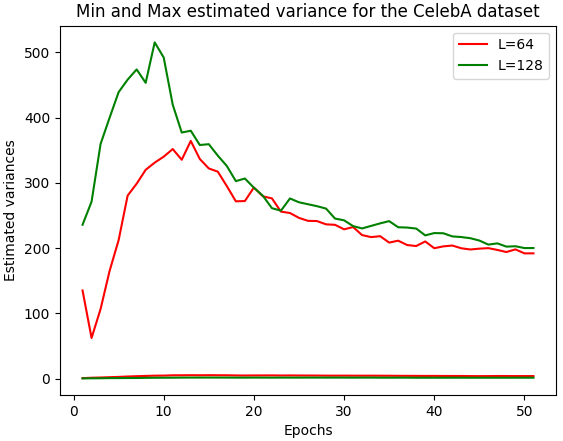}}
    \caption{CelebA}
  \end{subfigure}
  \begin{subfigure}{0.32\linewidth}
    \centerline{\includegraphics[width=0.95\textwidth]{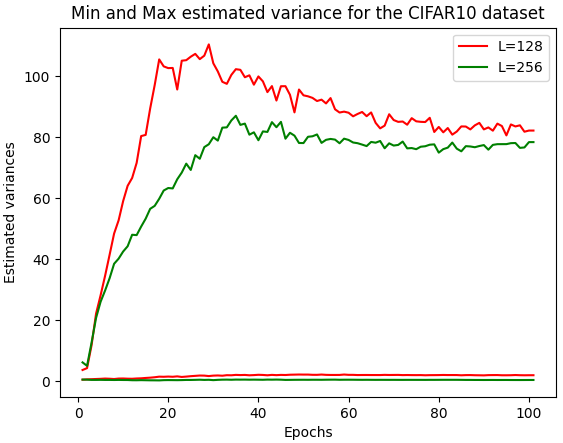}}
    \caption{CIFAR10}
  \end{subfigure}
  \caption{The minimum and maximum variances estimated by the ARD-VAE while training on the MNIST, CelebA and CIFAR10 datasets for multiple latent dimensions. The maximum estimated variances are orders of magnitude higher than the minimum estimated variances.}
  \label{fig:Est_Var_MNIST_CelebA_CIFAR10}
\end{figure*}

\begin{figure*}
\begin{center}
\centerline{\includegraphics[width=1.0\textwidth]{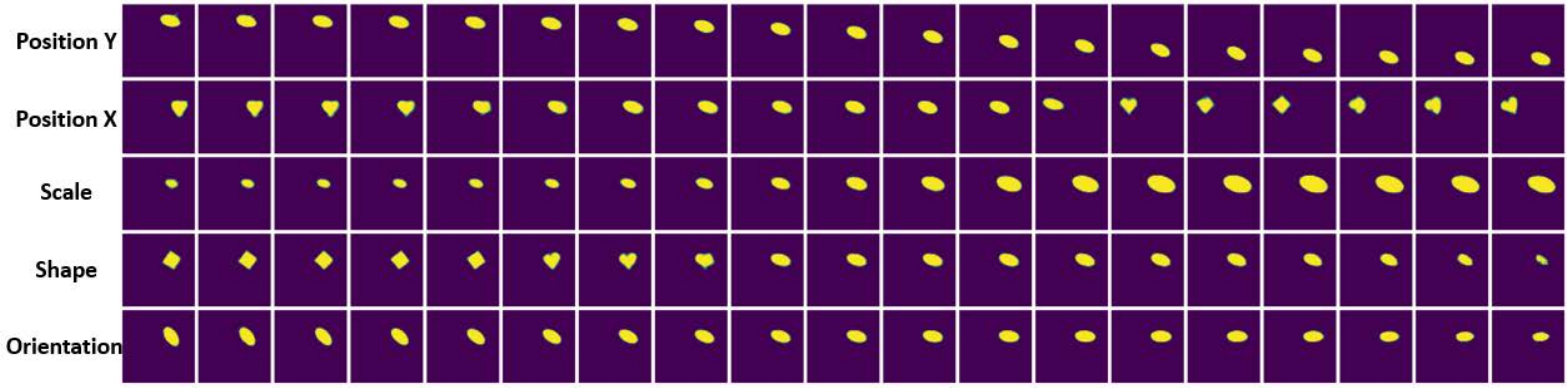}}
\caption{Latent traversal of the DSprites data set \cite{dsprites17} in the range $[-3\sigma, 3\sigma]$ using the relevant axes discovered by the ARD-VAE. The latent factors are mentioned in the left column. All latent factors are represented by independent latent axes with slight entanglement of \textbf{Shape} and \textbf{Position Y}.  The MIG score for this model is $\mathbf{0.35}$}
\label{img:ARD-VAE_DSprites}
\end{center}
\end{figure*}

\begin{table}
    \centering
    \resizebox{0.49\textwidth}{!}
    {
    \begin{tabular}{c c c c c c c}
    \hline
    {$\mid\mathcal{X{}}_{\boldsymbol\alpha}\mid$}
    &&\textsc{Active} &\textsc{FID}$\downarrow$ &\textsc{Precision}$\uparrow$ &\textsc{Recall}$\uparrow$ &\textsc{MSE}$\downarrow$ \\ [0.5ex] 
    \hline
    $10K$ 
    &&$152$ &$120.53$ &$0.56$ &$0.48$ &$0.005$ \\
    $20K$ 
    &&$153$ &$119.66$ &$0.53$ &$0.57$ &$0.005$ \\
    [1.1ex] \hline
    \end{tabular}
    }
    \caption{Effect of the size of the $\mid\mathcal{X{}}_{\boldsymbol\alpha}\mid$ on the performance of the ARD-VAE under different metrics for the \textbf{large ImageNet dataset}. We observe the ARD-AVE is resilient to the size of the $\mid\mathcal{X{}}_{\boldsymbol\alpha}\mid$ for a sufficiently large size, such as $\mid\mathcal{X{}}_{\boldsymbol\alpha}\mid=10K$. This observation is consistent with the results on the MNSIT and CIFAR10 datasets in \cref{tab:FID_Score_Gaussian_Samples}. Therefore, we demonstrate that the configuration of the ARD-VAE used in relatively smaller datasets, such as the MNIST and CIFAR10, seamlessly scales to the large ImageNet dataset.}\label{tab:ARD-VAE_ImageNet_Scalable_20K}
\end{table}

\begin{table}
    \centering
    \resizebox{0.49\textwidth}{!}
    {
    \begin{tabular}{c c c c c}
    \hline
    Dataset &$L\; (secs)$ &$2L\; (secs)$ &$4L\; (secs)$ \\ [0.5ex] 
    \hline
    MNIST $(L=16)$ &$0.36 \pm 0.02$ &$0.36 \pm 0.01$ &$0.36 \pm 0.01$ \\
    CIFAR10 $(L=128)$ &$0.64 \pm 0.02$ &$0.64 \pm 0.02$ &$0.65 \pm 0.02$ \\
    [1.1ex] \hline
    \end{tabular}
    }
    \caption{Time taken (in secs) in the computation of $\boldsymbol{b}_{L}$ on a 12GB NVIDIA TITAN V to get the updated $\boldsymbol{\hat{\sigma}}^2$ using $\mid\mathcal{X{}}_{\boldsymbol\alpha}\mid=10K$ (refer to the \cref{alg:ARD-VAE_Alg}). The inference time is indifferent to the size of the latent space, $L$. However, it increases with the complexity of the neural network, such as for the CIFAR10 dataset.} \label{tab:ARD-VAE_computation_time}
\end{table}

Considering the size of the training set ($50K$) of the MNIST and CIFAR10 datasets, $\mid\mathcal{X{}}_{\boldsymbol\alpha}\mid=10K$ is a large number. This motivated us to evaluate the performance of the ARD-VAE with different settings of $\mid\mathcal{X{}}_{\boldsymbol\alpha}\mid$ on the \emph{large} ImageNet dataset containing $\sim 1.28$ million training samples. We train the ARD-VAE on the ImageNet dataset with the configuration of $\mid\mathcal{X{}}_{\boldsymbol\alpha}\mid=\{10K, 20K\}$ and evaluate its performance under several metrics, such as the relevant dimensions (\textsc{Active}), FID score of the generated samples, precision-recall scores, and reconstruction loss (MSE). We have used the neural network architecture in \cref{tab:Encoder_decoder_NN_1} and followed the optimization strategies discussed in the section \ref{app:benchmark_arch}. From the results reported in \cref{tab:ARD-VAE_ImageNet_Scalable_20K}, we observe that the performance ARD seamlessly scales to the large ImageNet dataset, and its performance is not affected with more samples in $\mid\mathcal{X{}}_{\boldsymbol\alpha}\mid$. Therefore, we empirically demonstrate the robustness of the setting, $\mid\mathcal{X{}}_{\boldsymbol\alpha}\mid=10K$, across multiple datasets. We use $\mid\mathcal{X{}}_{\boldsymbol\alpha}\mid=10K$ for other large datasets (relative to the MNIST and CIFAR10 datasets) studied in this work, e.g., the CelebA ($\sim200K$), DSprites ($\sim700K$) and 3D Shapes ($\sim500K$) datasets.

\begin{figure*}
\begin{center}
\centerline{\includegraphics[width=1.0\textwidth]{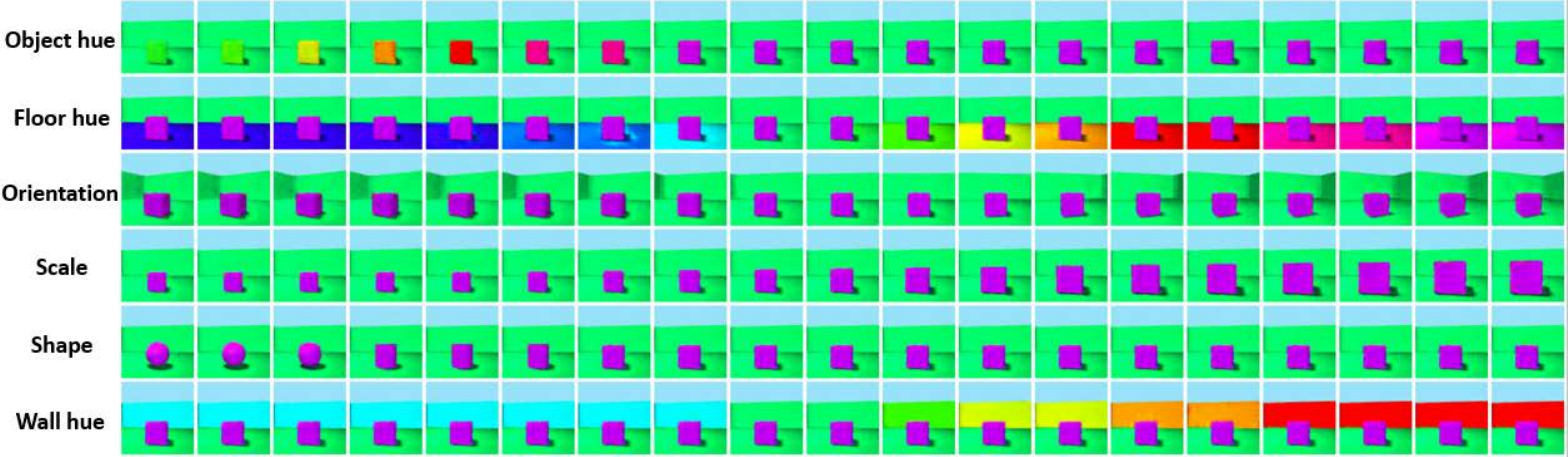}}
\caption{Latent traversal of the 3D Shapes data set \cite{3dshapes18} in the range  $[-3\sigma, 3\sigma]$ using the relevant axes discovered by the ARD-VAE. The latent factors are mentioned in the left column. All latent factors are represented by independent latent axes with almost no overlap between them. The MIG score for this model is $\mathbf{0.84}$}
\label{img:ARD-VAE_Shapes3D}
\end{center}
\end{figure*}

The parameters of the prior distribution, $p(\z{} \mid D{}_{\z{}})$, are updated every epoch (indicated by $u_{D_{\z{}}}$ in the \cref{alg:ARD-VAE_Alg}) using samples in the latent space. In \cref{tab:ARD-VAE_computation_time}, we report the time taken (in secs) on a 12GB NVIDIA TITAN V to estimate the parameters of the prior distribution using $\mid\mathcal{X{}}_{\boldsymbol\alpha}\mid=10K$ for the MNIST and CIFAR10 datasets. We observe the inference time is indifferent to the size of the latent space for both the datasets. The inference time increases with the complexity of the encoder-decoder architecture used for different datasets, such as the neural network used for the CIFAR10 dataset with more model parameters than the MNIST takes more time to estimate the parameters of the posterior distribution. However, as the parameters are updated \emph{only once} in an epoch, the inference time taken is negligible relative to the training duration. The time taken to estimate the distribution parameters for the ImageNet dataset is similar to the CIFAR10, as we use the same neural network architecture. Thus, the results in \cref{tab:ARD-VAE_computation_time}, demonstrates the feasibility of training the ARD-VAE on large datasets and bigger latent spaces. Moreover, we could successfully train the ARD-VAE on the ImageNet dataset on a single 12GB NVIDIA TITAN V using the architecture mentioned in \cref{tab:Encoder_decoder_NN_1} and parameters in \cref{tab:hyper-parameters}.

The prior distribution of the ARD-VAE is not fixed, unlike the regular VAE \cite{kingma2014auto, rezende2014stochastic}. Thus, we track the minimum and maximum estimated variances for the MNIST, CelebA and CIFAR10 datasets across the training epochs in \cref{fig:Est_Var_MNIST_CelebA_CIFAR10}. The minimum estimated variance is significantly less than the maximum variance across different experimental setup and the maximum estimated variance stabilizes after certain number of epochs, depending on the dataset. For all the datasets, the estimated variances (both the minimum and maximum) are similar for different sizes of the latent space. These figures illustrate the stability in the training of the ARD-VAE across multiple datasets.

\begin{figure*}
\begin{center}
\centerline{\includegraphics[width=1.0\textwidth]{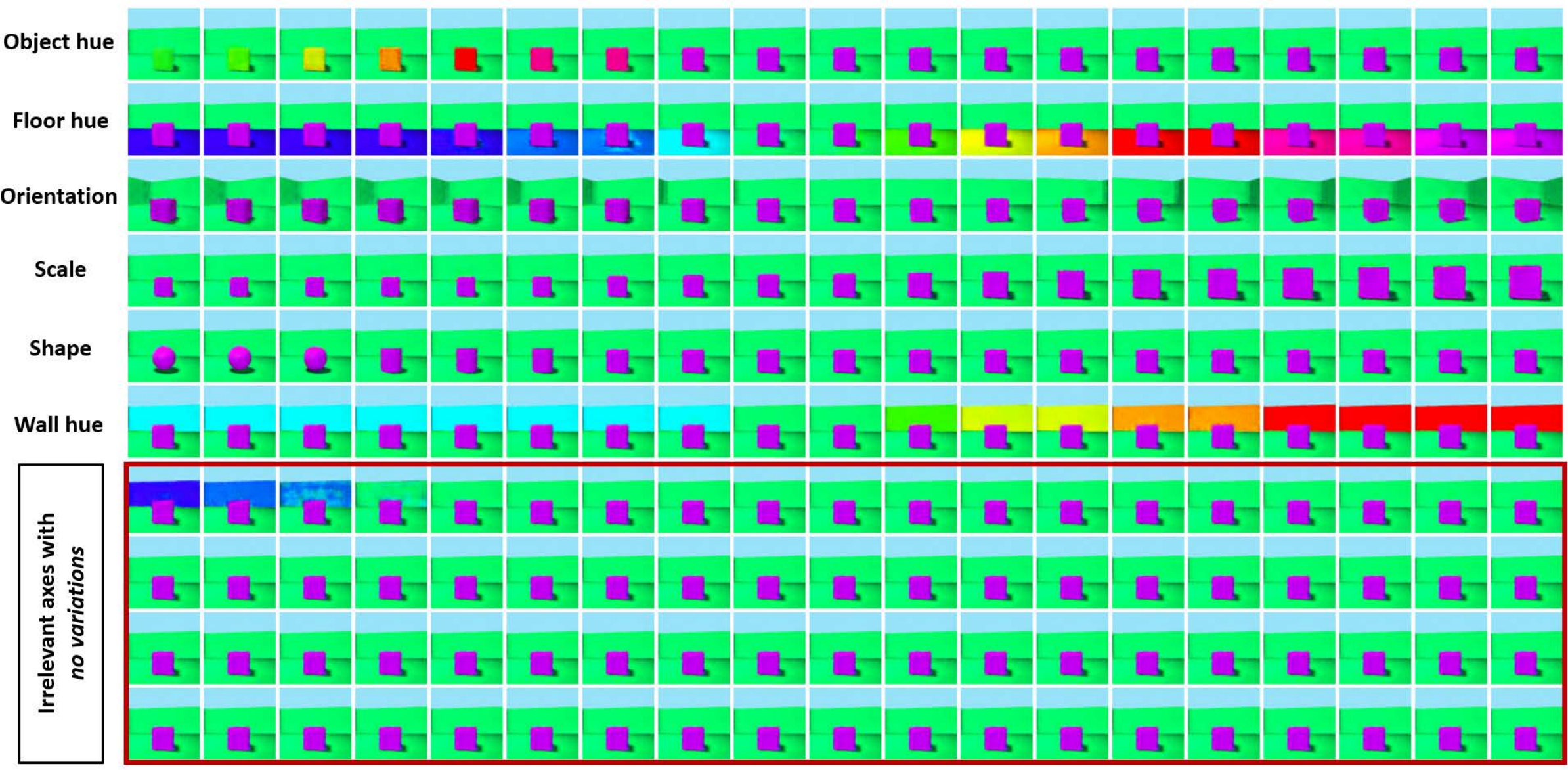}}
\caption{Latent traversal of the 3D Shapes data set \cite{3dshapes18} in the range  $[-3\sigma, 3\sigma]$ on all the latent axes, $L=10$, used in the training of the ARD-AVE on the 3D Shapes dataset, sorted by the relevance score proposed in the paper. The latent factors are mentioned in the left column for the relevant axes (total $6$), and additional axes (total $4$) are highlighted within the red bounding box that shows no variability in output in response to deviations along these axes. This elucidates our hypothesis about the behavior of \emph{irrelevant} latent axes.}
\label{img:invariance_3DShapes}
\end{center}
\end{figure*}

The ARD-VAE outperforms the competing methods under different disentanglement metrics on multiple datasets, as reported in Tab. 1 in the main paper. The information encoded by the relevant latent dimensions identified by the ARD-VAE when trained on the DSprites and 3D Shapes are shown in \cref{img:ARD-VAE_DSprites} and \cref{img:ARD-VAE_Shapes3D}, respectively. In both images, \cref{img:ARD-VAE_DSprites} and \cref{img:ARD-VAE_Shapes3D}, we traverse each relevant latent axis in the range $[-3\sigma, 3\sigma]$ to interpret the variability explained by the axes. From these images, we conclude that the ARD-VAE identifies all factors of variation present in both the DSprites and 3D Shapes datasets. In \cref{img:invariance_3DShapes}, we traverse all the latent axes, $L=10$, used in the training of the ARD-AVE on the 3D Shapes dataset, sorted by the relevance score proposed in the paper. We observe that the decoder produces no variability in output in response to the deviations along \emph{a few latent axes} with lower relevance score, highlighted within the red bounding box. This visualization demonstrates the presence of \emph{irrelevant} or \emph{superfluous} latent dimensions that the ARD-VAE correctly identifies.

\begin{table*}
\centering
\resizebox{\textwidth}{!}
{%
\begin{tabular}{c c c c c c c}
\hline
&&{MNIST}&&{CelebA}&&{CIFAR10\;\; \& \;\; ImageNet}\\ \hline
{Encoder: }
&&$\x{} \in \R{}^{32 \times 32 \times 1}$ 
&&$\x{} \in \R{}^{64 \times 64 \times 3}$
&& $\x{} \in \R{}^{32 \times 32 \times 3}$ \\
&&$\textsc{Conv64} \rightarrow \textsc{BN} \rightarrow \textsc{ReLU}$ 
&&$\textsc{Conv64} \rightarrow \textsc{BN} \rightarrow \textsc{ReLU}$
&& $\textsc{Conv128} \rightarrow \textsc{BN} \rightarrow \textsc{ReLU}$ \\
&&$\textsc{Conv128} \rightarrow \textsc{BN} \rightarrow \textsc{ReLU}$ 
&&$\textsc{Conv128} \rightarrow \textsc{BN} \rightarrow \textsc{ReLU}$
&& $\textsc{Conv256} \rightarrow \textsc{BN} \rightarrow \textsc{ReLU}$ \\
&&$\textsc{Conv256} \rightarrow \textsc{BN} \rightarrow \textsc{ReLU}$ 
&&$\textsc{Conv256} \rightarrow \textsc{BN} \rightarrow \textsc{ReLU}$
&& $\textsc{Conv512} \rightarrow \textsc{BN} \rightarrow \textsc{ReLU}$ \\
&&$\textsc{Conv512} \rightarrow \textsc{BN} \rightarrow \textsc{ReLU}$ 
&&$\textsc{Conv512} \rightarrow \textsc{BN} \rightarrow \textsc{ReLU}$
&& $\textsc{Conv1024} \rightarrow \textsc{BN} \rightarrow \textsc{ReLU}$ \\
&&$\textsc{Flatten}_{2 \times 2 \times 512} \rightarrow \textsc{FC}_{k \times 16} \rightarrow \textsc{None}$ &&$\textsc{Flatten}_{4 \times 4 \times 512} \rightarrow \textsc{FC}_{k \times 64} \rightarrow \textsc{None}$
&& $\textsc{Flatten}_{2 \times 2 \times 1024} \rightarrow \textsc{FC}_{k \times 128} \rightarrow \textsc{None}$ \\ \hline
{Decoder: }&&$\z{} \in \R{}^{16} \rightarrow \textsc{FC}_{2 \times 2 \times 512}$ &&$\z{} \in \R{}^{64}  \rightarrow \textsc{FC}_{8 \times 8 \times 512}$&& $\z{} \in \R{}^{128}  \rightarrow \textsc{FC}_{8 \times 8 \times 1024}$ \\
&&$\textsc{TransConv256} \rightarrow \textsc{BN} \rightarrow \textsc{ReLU}$ &&$\textsc{TransConv256} \rightarrow \textsc{BN} \rightarrow \textsc{ReLU}$&& $\textsc{TransConv512} \rightarrow \textsc{BN} \rightarrow \textsc{ReLU}$ \\
&&$\textsc{TransConv128} \rightarrow \textsc{BN} \rightarrow \textsc{ReLU}$ &&$\textsc{TransConv128} \rightarrow \textsc{BN} \rightarrow \textsc{ReLU}$&& $\textsc{TransConv256} \rightarrow \textsc{BN} \rightarrow \textsc{ReLU}$ \\
&&$\textsc{TransConv64} \rightarrow \textsc{BN} \rightarrow \textsc{ReLU}$ &&$\textsc{TransConv64} \rightarrow \textsc{BN} \rightarrow \textsc{ReLU}$&& $\textsc{TransConv3} \rightarrow \textsc{Sigmoid}$ \\
&&$\textsc{TransConv1} \rightarrow \textsc{Sigmoid}$ &&$\textsc{TransConv3} \rightarrow \textsc{Tanh}$&& \\
\hline
\end{tabular}
}
\caption{Encoder and decoder architectures used by all the methods for the MNIST, CelebA and CIFAR10 datasets \cite{tolstikhin2017wasserstein,RAE}.} 
\label{tab:Encoder_decoder_NN_1}
\end{table*}

\section{Experimental Settings}\label{app:benchmark_arch}
In the neural network architectures reported in \cref{tab:Encoder_decoder_NN_1}, \cref{tab:Encoder_decoder_NN_2}, \textsc{Conv}${n}$ and \textsc{TransConv}${n}$ define convolution and transpose convolution operation, respectively, with $n$ filters in the output. We have used $4 \times 4$ filters for all the datasets. The transpose convolution filters use a stride size of $2$ except for the last layer of the decoders used in the CelebA, CIFAR10 and ImageNet datasets. We represent the fully connected layers as $\textsc{FC}_{k \times n}$ with $k \times n$ nodes, where $k=1$ for all the methods, except the VAE and $\beta$-TCVAE that use $k=2$. Activation functions used in the networks are ReLU (\textsc{ReLU}), Leaky ReLU (\textsc{LReLU}), sigmoid (\textsc{Sigmoid}), and hyperbolic tangent (\textsc{Tanh}). Input is in the range $[0, 1]$ for all the datasets except CelebA, for which the input is mapped to the range $[-1, 1]$. The encoder-decoder architectures of the MaskAAE \cite{vae_prune_mask_AAE_reg_2020} and GECO-$L_0$-ARM-VAE \cite{vae_prune_L0_reg_2021} are obtained from the respective papers.

\begin{table*}
\centering
{%
\begin{tabular}{c c c c c}
\hline
&&{DSprites}&&{3D Shapes}\\ \hline
{Encoder: }
&&$\x{} \in \R{}^{64 \times 64 \times 1}$ 
&&$\x{} \in \R{}^{64 \times 64 \times 3}$ \\
&&$\textsc{Conv32} \rightarrow \textsc{ReLU}$ 
&&$\textsc{Conv32} \rightarrow \textsc{ReLU}$ \\
&&$\textsc{Conv32} \rightarrow \textsc{ReLU}$ 
&&$\textsc{Conv32} \rightarrow \textsc{ReLU}$ \\
&&$\textsc{Conv64} \rightarrow \textsc{ReLU}$ 
&&$\textsc{Conv64} \rightarrow \textsc{ReLU}$ \\
&&$\textsc{Conv64} \rightarrow \textsc{ReLU}$ 
&&$\textsc{Conv64} \rightarrow \textsc{ReLU}$ \\
&&$\textsc{Flatten}_{4 \times 4 \times 64} \rightarrow \textsc{FC}_{k \times 6} \rightarrow \textsc{None}$ &&$\textsc{Flatten}_{4 \times 4 \times 64} \rightarrow \textsc{FC}_{k \times 6} \rightarrow \textsc{None}$ \\ \hline
{Decoder: }
&&$\z{} \in \R{}^{6} \rightarrow \textsc{FC}_{4 \times 4 \times 64}$ 
&&$\z{} \in \R{}^{6} \rightarrow \textsc{FC}_{4 \times 4 \times 64}$ \\
&&$\textsc{TransConv64} \rightarrow \textsc{ReLU}$ 
&&$\textsc{TransConv64} \rightarrow \textsc{ReLU}$ \\
&&$\textsc{TransConv32} \rightarrow \textsc{ReLU}$ 
&&$\textsc{TransConv32} \rightarrow \textsc{ReLU}$ \\
&&$\textsc{TransConv32} \rightarrow \textsc{ReLU}$ 
&&$\textsc{TransConv32} \rightarrow \textsc{ReLU}$ \\
&&$\textsc{TransConv1} \rightarrow \textsc{None}$ 
&&$\textsc{TransConv3} \rightarrow \textsc{Sigmoid}$ \\
\hline
\end{tabular}
}
\caption{Encoder and decoder architectures used by all the methods for the DSprites and 3D Shapes datasets \cite{Factor_VAE_ICML_2018,Dis_benchmark_ICML_2019}.} \label{tab:Encoder_decoder_NN_2}
\end{table*}

\begin{table*}
\centering
\resizebox{\textwidth}{!}
{%
\begin{tabular}{c c c c c c c c c c c c c c c}
\hline
Method&&Parameters&&{MNIST}&&{CelebA}&&{CIFAR10}&&{ImageNet}&&{DSprites}&&{3D Shapes}\\ \hline
\textsc{$\beta$-TCVAE} && $\beta$: && $2$ && $2$ && $2$ && $2$  && $5$ && $5$\\
\textsc{DIP-VAE-I} && $(\lambda_{od}, \lambda_{d})$: && NA && NA && NA && NA  && $(10, 100)$ && $(10, 100)$ \\
\textsc{DIP-VAE-II} && $(\lambda_{od}, \lambda_{d})$: && NA && NA && NA && NA && $(10, 10)$ && $(10, 10)$ \\
\textsc{WAE} && \textsc{Recons-scalar}: && $0.05$ && $0.05$ && $0.05$ && $0.05$  && NA && NA\\
\textsc{WAE} && $\beta$: && $10$ && $100$ && $100$ && $100$ && NA && NA \\
\textsc{RAE} && $\beta$: && $1e-04$ && $1e-04$ && $1e-03$ && $1e-03$ && $1e-04$ && $1e-04$ \\
\textsc{RAE} && \textsc{Dec-L}$2$-\textsc{reg}: && $1e-07$ && $1e-07$ && $1e-06$ && $1e-06$ && $1e-06$ && $1e-06$\\
\textsc{GECO-$L_0$-ARM-VAE} && $\tau$: && $10.0$ && $300.0$ && $25.0$ && $15.0$  && NA && NA\\
\textsc{ARD-VAE} && $\beta$: && $0.5$ && $1.0$ && $0.05$ && $0.05$  && $5.0$ && $5.0$\\
\hline
\end{tabular}
}
\caption{Optimization settings for different methods.} 
\label{tab:hyper-parameters}
\end{table*}

\begin{table}
    \resizebox{0.49\textwidth}{!}
    {
    \begin{tabular}{c c c c c c c}
    \hline
    \multicolumn{3}{c}{DSprites}&&\multicolumn{3}{c}{3D Shapes}\\ \cline{1-3}\cline{5-7}
\textsc{Initial} &\textsc{Active} &\textsc{GT}
&&\textsc{Initial} &\textsc{Active} &\textsc{GT}\\ \hline
    $10$ &$5.80 \pm 0.40$ &$6$
    &&$10$ &$6.40 \pm 0.49$ &$6$\\
    
    $15$ &$6.00 \pm 0.00$ &$6$
    &&$15$ &$6.40 \pm 0.49$ &$6$ \\
    
    $20$ &$5.80 \pm 0.40$ &$6$
    &&$20$ &$6.40 \pm 0.49$ &$6$\\
    
    $30$ &$5.80 \pm 0.40$ &$6$
    &&$30$ &$6.20 \pm 0.40$ &$6$\\
    \hline
    \end{tabular}
    }
    \caption{The number of the active dimensions (\textsc{Active}) estimated by the ARD-VAE matches closely with the ground truth (\textsc{GT}).} \label{tab:Synthetic_Active}
\end{table}

\begin{table*}
    \centering
    \resizebox{\textwidth}{!}
    {%
    \begin{tabular}{c c c c c c c c c}
    \hline
    \multirow{2}{*}{\textsc{Method}}&&\multicolumn{3}{c}{\textsc{DSprites}}&&\multicolumn{3}{c}{\textsc{3D Shapes}}\\ 
    \cline{3-6}\cline{7-9}
    &&\textsc{FactorVAE metric}\;\;$\uparrow$&\textsc{MIG}\;\;$\uparrow$ &\textsc{Active}&&\textsc{FactorVAE metric}\;\;$\uparrow$&\textsc{MIG}\;\;$\uparrow$&\textsc{Active}\\ \hline
    $\beta$-TCVAE $(L=10)$ 
    && $\mathbf{81.15}$ & ${0.23}$ & $10$
    && ${84.29}$ & ${0.48}$ & $10$ \\
    ARD-VAE $(L=10)$ 
    && ${66.63}$ & $\mathbf{0.26}$ & $6$
    && $\mathbf{91.40}$ & $\mathbf{0.84}$ & $6$\\  
    \hline
    \end{tabular}
    }
    \caption{Disentanglement scores of competing methods, where we initialize the model parameters of the different methods with the \emph{same} seed for multiple datasets. In this experiment, we train the $\beta$-TCVAE (a baseline method comparable to the ARD-VAE in Tab. 1 in the main paper) with the latent dimensions $L=10$ and compare its performance with the ARD-VAE under the same setting. The \textsc{Active} indicates the number of the latent dimensions used by different methods to compute the metric scores (higher is better). We train the ARD-VAE with $L=10$ and find the number of relevant dimensions as $6$ (\textsc{Active}). The \textbf{best} score is in \textbf{bold}. The ARD-VAE mostly outperforms the $\beta$-TCVAE (similar to Tab. 1 in the main paper), even with $L=10$.} \label{tab:Disentanglement_study_Baseline_Ablation}
\end{table*}

\begin{table*}
    \centering
    \resizebox{\textwidth}{!}
    {
    \begin{threeparttable}
    \begin{tabular}{c c c c c c c c c c c}
    \hline
    &&\multicolumn{4}{c}{CelebA\;\;$(L=64)$} 
    &&\multicolumn{4}{c}{ImageNet\;\;$(L=256)$}\\ 
    \cline{3-6} \cline{8-11}
    &&\textsc{Active} &\textsc{FID}$\downarrow$ &\textsc{Precision}$\uparrow$ &\textsc{Recall}$\uparrow$
    &&\textsc{Active} &\textsc{FID}$\downarrow$ &\textsc{Precision}$\uparrow$ &\textsc{Recall}$\uparrow$ 
    \\ \hline
    VAE
    && $64$ &$\underline{49.89 \pm 0.57}$ & $0.79 \pm 0.03$ & $\underline{0.75 \pm 0.03}$
    && $256$ &$180.44 \pm 0.69$ & $0.12 \pm 0.01$ & $0.31 \pm 0.04$ \\
    $\beta$-TCVAE 
    &&  $64$  &$50.14 \pm 0.78$ & $0.78 \pm 0.02$ & $0.70 \pm 0.05$ 
    &&  $256$  &$226.27 \pm 0.48$ & $0.04 \pm 0.01$ & $0.25 \pm 0.01$ \\
    RAE 
    &&  $64$  &$\mathbf{48.81 \pm 1.02}$ & $\underline{0.81 \pm 0.02}$ & $\mathbf{0.77 \pm 0.04}$
    &&  $256$  &$226.70 \pm 30.14$ & $0.08 \pm 0.05$ & $0.11 \pm 0.05$ \\
    WAE 
    &&  $64$  &$72.01 \pm 2.26$ & $0.64 \pm 0.05$ & $\underline{0.75 \pm 0.02}$
    &&  $256$  &$278.32 \pm 5.83$ & $0.03 \pm 0.00$ & $0.04 \pm 0.08$ \\
    GECO-$L_0$-ARM-VAE
    &&$35.60 \pm 3.07$ &$294.97 \pm 28.45$ &$0.00 \pm 0.00$ &$0.00 \pm 0.00$
    &&$124.00 \pm 9.34$ &$\underline{177.40 \pm 40.21}$ &$\underline{0.27 \pm 0.09}$ &$\underline{0.26 \pm 0.10}$\\
    MaskAAE
    &&$5.4 \pm 0.80$ & $333.40 \pm 10.91$ & $0.01 \pm 0.02$ & $0.00 \pm 0.00$
    &&$0.0 \pm 0.00$\tnote{*} &\textemdash &\textemdash &\textemdash \\
    ARD-VAE 
    && $53.40 \pm 0.49$ & ${50.73 \pm 0.29}$ & $\mathbf{0.85 \pm 0.02}$ & $0.73 \pm 0.02$    
    && $152.0 \pm 1.1$ &$\mathbf{121.21 \pm 1.16}$ &$\mathbf{0.54 \pm 0.02}$ &$\mathbf{0.51 \pm 0.03}$ \\
    [1.1ex] \hline
    \end{tabular}
    \begin{tablenotes}
    \item[*] The MaskAAE collapsed all the dimensions for the ImageNet dataset after $30$ epochs. Thus, we skip the evaluation of the MaskAAE for the ImageNet dataset.
    \end{tablenotes}
    \end{threeparttable}
    }
    \caption{The FID and precision-recall scores of the generated samples, along with the number of latent dimensions (\textsc{Active}) used by the competing methods to compute different metric scores for the CelebA and ImageNet datasets. The \textbf{best} score under a metric is in \textbf{bold} and the \underline{second best} score is \underline{{underlined}}. The ARD-VAE outperforms other methods \emph{by far} on the ImageNet dataset and is comparable to the competing methods on the CelebA dataset.} \label{tab:FID_Score_complete_CelebA_ImageNet}
\end{table*}

\begin{table*}
    \centering
    \resizebox{\textwidth}{!}
    {%
    \begin{threeparttable}
    \begin{tabular}{|c||c||c||c||c|}\hline
    Method & MNIST $(L=16) \downarrow$ 
    & CelebA $(L=64) \downarrow$ 
    & CIFAR10  $(L=128) \downarrow$ 
    & ImageNet  $(L=256) \downarrow$
    \\ [0.5ex] 
    \hline\hline
    VAE             & $0.012 \pm 0.000$ & $\underline{0.021 \pm 0.000}$ & $0.016 \pm 0.000$ &$0.018 \pm 0.000 $\\
    $\beta$-TCVAE   & $0.018 \pm 0.000$ & $0.024 \pm 0.000$ & $0.021 \pm 0.000 $ &$0.022 \pm 0.000$ \\
    RAE             & $\mathbf{0.003 \pm 0.000}$ & $\mathbf{0.020 \pm 0.000}$ & $\mathbf{0.006 \pm 0.000}$ &$\mathbf{0.003 \pm 0.000}$\\
    WAE             &$\underline{0.004 \pm 0.000}$ & $\mathbf{0.020 \pm 0.000}$ & $\underline{0.007 \pm 0.000}$ &$0.007 \pm 0.000$\\
    GECO-$L_0$-ARM-VAE & $\underline{0.004 \pm 0.001}$ & $0.029 \pm 0.002$ & $0.011 \pm 0.001$ &$0.006 \pm 0.000$\\
    MaskAAE & $0.091 \pm 0.091$ & $0.093 \pm 0.093$ & $0.199 \pm 0.199$ &\textemdash\tnote{*} \\
    ARD-VAE & ${0.008 \pm 0.000}$ & $\underline{0.021 \pm 0.000}$ & $\mathbf{0.006 \pm 0.000}$ &$\underline{0.005 \pm 0.000}$\\ [1.5ex] \hline
    \end{tabular}
    \begin{tablenotes}
    \item[*] The MaskAAE collapsed all the dimensions for the CIFAR10 dataset after $20$ epochs. Thus, we skip the evaluation of the MaskAAE for the CIFAR10 dataset.
    \end{tablenotes}
    \end{threeparttable}
    }
    \caption{MSE per pixel of the competing methods (averaged over $5$ different runs) on the benchmark datasets (lower is better). The \textbf{best} score is in \textbf{bold}, and the \underline{second best} score is \underline{{underlined}}. The number of \textsc{active} dimensions of the GECO-$L_0$-ARM-VAE, MaskAAE and ARD-VAE (refer to \cref{tab:FID_Score_complete_Rel_Axes}) are used for computing the MSE. However, for the remaining methods we use all the initial $(L)$ dimensions.} \label{tab:mse}
\end{table*}

\begin{table*}
    \centering
    \resizebox{\textwidth}{!}
    {
    \begin{tabular}{c c c c c c c c c c c c c}
    \hline
    \multirow{2}{*}{\textsc{Method}}    
    &&\multicolumn{3}{c}{MNIST\;\;$(L=16)$}
    &&\multicolumn{3}{c}{CIFAR10\;\;$(L=128)$}
    &&\multicolumn{3}{c}{ImageNet\;\;$(L=256)$} \\
    \cline{3-5}\cline{7-9}\cline{11-13}
    &&\textsc{Active} &ARD-VAE-\textsc{All} &ARD-VAE
    &&\textsc{Active} &ARD-VAE-\textsc{All} &ARD-VAE
    &&\textsc{Active} &ARD-VAE-\textsc{All} &ARD-VAE \\ \hline
    $L$
    && $12.8 \pm 0.40$ &$21.94 \pm 0.67$ & $22.24 \pm 0.57$
    && $105.8 \pm 1.33$ &$85.05 \pm 0.84$ & $87.56 \pm 1.21$
    && $152.0 \pm 1.1$ &$107.9 \pm 1.08$ & $121.21 \pm 1.16$ \\
    $2L$
    && $12.6 \pm 0.49$ &$21.97 \pm 0.23$ & $22.30 \pm 0.33$
    && $116.4 \pm 3.72$ &$83.73 \pm 2.76$ & $86.50 \pm 1.43$
    && $163.0 \pm 1.41$ &$107.02 \pm 1.34$ & $123.34 \pm 1.11$ \\
    $4L$
    && $12.4 \pm 0.49$ &$22.30 \pm 0.48$ & $22.31 \pm 0.58$
    && $117.8 \pm 15.04$ &$84.92 \pm 0.92$ & $87.88 \pm 1.80$
    && $161.0 ±\pm 2.28$ &$107.78 \pm 0.64$ & $123.30 \pm 0.51$ \\
    [1.5ex] \hline
    \end{tabular}
    }
    \caption{In this experiment, we assess the impact of the \emph{pruning} latent dimensions in the ARD-VAE by comparing the FID scores over multiple datasets. We compare the FID scores (of the generated samples) of the proposed ARD-VAE with its variant using all the latent axes, referred to herein as ARD-VAE-\textsc{All}, to produce the metric scores. The number of relevant latent dimensions the ARD-VAE uses are reported as \textsc{Active}. The ARD-VAE produces consistent estimates of the \textsc{Active} dimensions and FID scores for different values of $L$.}
    \label{tab:Information_loss_ARD_Different_L}
\end{table*}

\begin{table*}
    \centering
    \resizebox{\textwidth}{!}
    {
    \begin{tabular}{c c c c c c c c c c c c c c}
    \hline
    \multirow{2}{*}{\textsc{Method}}
    &&\multicolumn{2}{c}{MNIST\;\;$(L=16)$}
    &&\multicolumn{2}{c}{CelebA\;\;$(L=64) $}
    &&\multicolumn{2}{c}{CIFAR10\;\;$(L=128)$}
    &&\multicolumn{2}{c}{ImageNet\;\;$(L=256)$} \\ 
    \cline{3-4}\cline{6-7}\cline{9-10}\cline{12-13}
    &&\textsc{Active} &\textsc{FID}$\downarrow$
    &&\textsc{Active} &\textsc{FID}$\downarrow$
    &&\textsc{Active} &\textsc{FID}$\downarrow$
    &&\textsc{Active} &\textsc{FID} $\downarrow$ 
    \\ \hline
    \parbox{1.5in}{\centering \small  \textit{\textbf{Best}} method among \\ the baseline methods}
    && $16$ & $\mathbf{18.79 \pm 0.31}$
    && $64$  &$\mathbf{48.81 \pm 1.02}$
    && $128$  & ${94.34 \pm 1.58}$
    && $124.00 \pm 9.34$ &${177.40 \pm 40.21}$ \\ [1.2ex]
    ARD-VAE 
    && $12.80 \pm 0.40$ & $22.24 \pm 0.57$ 
    && $53.40 \pm 0.49$ & ${50.73 \pm 0.29}$ 
    && $105.80 \pm 1.33$ & $\mathbf{87.56 \pm 1.21}$ 
    && $152.0 \pm 1.10$ & $\mathbf{121.21 \pm 1.16}$ \\ 
    [1.2ex] \hline
    \end{tabular}
    }
    \caption{In this analysis, we assess the information loss due to the pruning of latent dimensions in the ARD-VAE by comparing its FID scores (of the generated samples) with the best method among the baseline methods (studied in this work) over multiple datasets. The number of latent dimensions used by competing methods are reported as \textsc{Active}. The \textbf{best} FID score is in \textbf{bold}.}
    \label{tab:Information_loss_ARD_Best}
\end{table*}

\begin{table*}
    \centering
    \resizebox{\textwidth}{!}
    {
    \begin{tabular}{c c c c c c c c c}
    \hline
    &\multicolumn{2}{c}{MNIST\;\;$(L=16)$}&&\multicolumn{2}{c}{CelebA\;\;$(L=64) $}&&\multicolumn{2}{c}{CIFAR10\;\;$(L=128)$}\\ \cline{2-3}\cline{5-6}\cline{8-9}
    &\textsc{Active} &\textsc{FID}$\downarrow$&
    &\textsc{Active} &\textsc{FID}$\downarrow$&
    &\textsc{Active} &\textsc{FID} $\downarrow$\\ \hline
    VAE \
    & $10.20 \pm 0.40$ & $29.33 \pm 0.41$
    &&  $55.20 \pm 0.40$ & ${50.79 \pm 0.64}$
    &&  $26.40 \pm 0.49$ & $155.13 \pm 1.45$ \\
    $\beta$-TCVAE 
    &  $7.40 \pm 0.49$ & $ 51.35 \pm 0.95 $
    && $51.40 \pm 0.49$ & $51.61 \pm 0.76$
    &&  $16.20 \pm 0.40$ & $186.03 \pm 1.82$ \\
    RAE 
    &  $16.00 \pm 0.00$ & $\mathbf{18.85 \pm 0.43}$
    && $63.00 \pm 0.00$ & $\mathbf{49.06 \pm 1.16}$
    &&  $125.00 \pm 0.00$ & $\underline{92.66 \pm 1.61}$ \\
    WAE 
    &  $16.00 \pm 0.00$ & $24.78 \pm 0.77$
    && $63.00 \pm 0.00$ & $61.18 \pm 1.60$
    &&  $127.00 \pm 0.00$ & $136.79 \pm 1.35$ \\
    GECO-$L_0$-ARM-VAE 
    & $10.00 \pm 1.10$ & $304.75 \pm 64.29$
    &&$35.60 \pm 3.07$ & $294.97 \pm 28.45$
    &&$68.00 \pm 2.97$ & $320.75 \pm 45.09$ \\
    MaskAAE
    &$9.80 \pm 1.60$ & $144.92 \pm 16.80$
    &&$5.4 \pm 0.80$ & $333.40 \pm 10.91$
    &&$3.80 \pm 0.40$ & $298.30 \pm 8.44$ \\
    ARD-VAE 
    & $12.80 \pm 0.40$ & $\underline{22.24 \pm 0.57}$
    && $53.40 \pm 0.49$ & $\underline{50.73 \pm 0.29}$
    && $105.80 \pm 1.33$ & $\mathbf{87.56 \pm 1.21}$ \\ [1.5ex]
    \hline
    \end{tabular}
    }
    \caption{The FID scores of the generated samples along with the number of (\textsc{Active}) latent dimensions used by the competing methods. The \textbf{best} FID score is in \textbf{bold} and the \underline{second best} is \underline{{underlined}}. The \textsc{Active} dimensions change for the VAE, $\beta$-TCVAE, RAE and WAE.}
    \label{tab:FID_Score_complete_Rel_Axes}
\end{table*}

We use the Adam optimizer in all experiments (learning rate set to $5e-04$) with a learning rate scheduler (ReduceLROnPlateau) that reduces the learning rate by $0.5$ if the validation loss does not improve for a maximum of $10$ epochs except the MaskAAE and GECO-$L_0$-ARM-VAE due to their sensitivity in optimization of the trainable parameters. Moreover, the MaskAAE uses a specific training recipe. All the methods are trained for $50$, $50$, $100$, and $50$ epochs for the MNIST, CelebA, CIFAR10, and ImageNet datasets, respectively, with a few exceptions for the MNIST dataset. The VAE, $\beta$-TCVAE, GECO-$L_0$-ARM-VAE, and MaskAEE are trained for 100 epochs for the MNIST dataset as it improved the model performance. In the disentanglement analysis, all the methods are trained for $35$ and $60$ epochs \cite{Dis_benchmark_ICML_2019} for the DSprites and 3D Shapes datasets, respectively.

We use a batch size of $100$ for training all the methods, except the MaskAAE, which is trained using a batch size of $64$ (to maintain consistency with their implementation). All the hyperparameters of the MaskAAE are set according to the GitHub repo in \url{https://github.com/arnabkmondal/MaskAAE} for all the datasets. We have tuned the hyperparameters of the MaskAAE over a series (i.e., dozens) of experiments, and no better results could be obtained. For the GECO-$L_0$-ARM-VAE, we have used the code shared by the authors \cite{vae_prune_L0_reg_2021}. Only the target reconstruction loss ($\tau$) in GECO-$L_0$-ARM-VAE was the hyperparameter in our analysis. For the DIP-VAE-I and DIP-VAE-II, we have followed the suggested hyperparameters in the paper \cite{DIP_VAE_ICLR_2018}. For the $\beta$-TCVAE, we have empirically determined the strength of the regularization loss for different data sets as we could not find them in the literature. We set $\beta=2$ for the MNIST, CelebA, and CIFAR10 data sets as higher $\beta$ resulted in poor reconstruction. Table \ref{tab:hyper-parameters} reports the specific hyperparameters for different methods. The $u_{D_{\z{}}}$ in the algorithm \ref{alg:ARD-VAE_Alg} is set to $1$ and $\mid\mathcal{X{}}_{\boldsymbol\alpha}\mid=10K$ for all the datasets studied in this work.

\section{Results}\label{app:results}
In this section, we report the results on the CelebA dataset for all the competing methods, the reconstruction loss on different datasets, an ablation study on the effect of the relevance score estimation method on the performance of the ARD-VAE, and demonstrate the use of the relevance score estimation method in the determination of the number of relevant/active dimensions for other competing methods studied in this work.

\textbf{Results on the synthetic dataset : } We know the number of latent factors used in the generation of the synthetic datasets, the DSprites \cite{dsprites17} and 3D Shapes \cite{3dshapes18}, which is $6$ for both datasets. Thus, the number of \emph{known} latent factors in these datasets serves as the \emph{ground truth} for the number of relevant (or \textsc{active}) dimensions estimated by the ARD-VAE. In this experiment, we set the initial size of the latent space, $L={10,15,20,30}$, and train the ARD-VAE to estimate the number of relevant axes for both datasets. From the results reported in \cref{tab:Synthetic_Active}, we observe the number of active dimensions estimated by the ARD-VAE closely matches the ground truth for different initial bottleneck dimensions.

In the disentanglement analysis of the synthetic datasets (discussed in the main paper), we leverage the 
information of the number of the {\em known latent factors} of the DSprites \cite{dsprites17} and Shapes3D \cite{3dshapes18} dataset to set the size of the latent space ($L=6$) for the baseline methods. However, we train the ARD-VAE with some additional latent dimensions ($L=10$), such that it identifies the true generative factors of the datasets and discards the unnecessary axes from the set of relevant axes. For a fair comparison, we train the $\beta$-TCVAE, a baseline method comparable to the ARD-VAE in Tab. 1 in the main paper, with the latent dimensions $L=10$ and compare its performance with the ARD-VAE under the same setting. The result of this ablation study is reported in \cref{tab:Disentanglement_study_Baseline_Ablation}. Even under this setting, the performance of the ARD-AVE is better than the $\beta$-TCVAE under all the scenarios but the FactorVAE metric on the DSprites dataset.

\textbf{Results on the CelebA and ImageNet dataset : } In this analysis, we report the FID \cite{IS_FID_1} and precision-recall \cite{precision_recall_distributions} scores of the competing methods in \cref{tab:FID_Score_complete_CelebA_ImageNet}. For the CelebA dataset, the performance of the competing methods are comparable with small variations. The ARD-VAE could achieve the performance of the RAE with $\sim$54 latent dimensions, which is $10$ dimensions less. Similar to the MNIST and CIFAR10 dataset, we could not train the GECO-$L_0$-ARM-VAE and MaskAAE on the CelebA dataset using the neural network architecture reported in \cref{tab:Encoder_decoder_NN_1}.

We evaluate the scalability and robustness of the ARD-AVE, and we train the ARD-VAE on the ImageNet dataset \cite{ImageNet}, i.e., a \emph{complex} and \emph{larger} dataset. In our analysis, we train the ARD-VAE along with the baseline methods on the ImageNet dataset with the image resolution of $32\times32$. The comparison of the proposed method with the competing methods under several evaluation metrics is reported in \cref{tab:FID_Score_complete_CelebA_ImageNet}. The ARD-AVE compares favorably to all other methods by a \emph{large margin} and \emph{prunes} more than $100$ latent dimensions to model the complex dataset. The MaskAAE collapsed all the latent dimensions on the ImageNet dataset when trained using the neural network architecture reported in \cref{tab:Encoder_decoder_NN_1}. The number of relevant dimensions estimated by the ARD-VAE are consistent even when trained with different sizes of the latent space, such as $L={512, 1024}$ (refer to \cref{tab:Information_loss_ARD_Different_L}). We could successfully train the ARD-VAE on the ImageNet
dataset on a single 12GB NVIDIA TITAN V  using the architecture mentioned in \cref{tab:Encoder_decoder_NN_1} and optimization parameters in \cref{tab:hyper-parameters}. The training for $50$ epochs with $\mid\mathcal{X{}}_{\boldsymbol\alpha}\mid=10K$ took around $\sim17$ hours. Therefore, the training of the ARD-VAE on the ImageNet dataset demonstrates the \emph{stability, robustness}, and \emph{effectiveness} of the proposed method in modeling a large dataset with a lot of variability.

\textbf{Reconstruction loss of the competing methods: } In \cref{tab:mse}, we report the mean-square error (MSE) loss per pixel on the MNIST, CelebA, CIFAR10, and ImageNet datasets. The number of \textsc{active} dimensions of the GECO-$L_0$-ARM-VAE, MaskAAE and ARD-VAE (refer to \cref{tab:FID_Score_complete_Rel_Axes} and \cref{tab:FID_Score_complete_CelebA_ImageNet}) are used for computing the MSE. However, for the remaining methods we use all the latent $(L)$ dimensions. The MaskAAE
collapsed all the latent dimensions on the ImageNet dataset, and,thus, we could not compute the MSE. Though the GECO-$L_0$-ARM-VAE has a comparable MSE loss, it failed to model the data distribution. taset. The choice of $\beta$ for the ARD-VAE produces comparable MSE, except the MNIST dataset. This is due to the lack of extensive hyperparameter tuning in the ARD-VAE. However, that did not impact the performance of the ARD-VAE significantly.

\textbf{Ablation study on the relevance score estimation: } In this work, we proposed a method to compute the relevancy of latent axes and used the same to estimate the number of active dimensions that are sufficient to model a data distribution. However, it is important to evaluate the performance of the ARD-VAE using all the latent axes. To this end, we compute the FID scores of the ARD-VAE on the MNIST,  CIFAR10, and the ImageNet datasets trained on latent spaces of different sizes for an individual dataset, shown in \cref{tab:Information_loss_ARD_Different_L}. The ARD-VAE that uses all the latent axes is dubbed as the ARD-VAE-\textsc{All}, i.e., it does not \emph{prunes} unnecessary/superfluous dimensions. The \textsc{Active} in \cref{tab:Information_loss_ARD_Different_L} indicates the number of relevant dimensions used by the ARD-VAE for the evaluation of the FID scores under different settings. This analysis serves a \emph{proxy} to the loss of information due to the \emph{pruning} of \emph{irrelevant} dimensions.

We observe the FID scores computed using the relevant axes (\textsc{Active}) are marginally higher than the scores computed using all the axes (\textsc{All}). However, we get rid of many redundant latent axes using the relevance score estimation technique. For e.g., the FID score for the CIFAR10 (slightly higher) with an initial dimension of $4L=512$ is produced using \emph{only} $\sim$118 latent axes. Similarly, for the ImageNet dataset, the ARD-VAE prunes $\sim 863$ irrelevant dimensions (estimated using the proposed relevance score) when trained with an initial dimension of $4L=1024$ \emph{without a significant change} in the FID score. This experiment illustrates that the modeling of the data distributions by the ARD-VAE using \emph{only} the relevant latent axes \emph{preserves most of the information} compared to the ARD-VAE-\textsc{All} using all latent axes, as indicated by a small degradation in the FID scores.

To better understand the loss of information due to the pruning of latent dimensions in the ARD-VAE, we compare the ARD-VAE with the \emph{best} performing baseline method determined in terms of the FID score. The best baseline method considers all the latent dimensions used in training the model. Comparing the FID scores reported in \cref{tab:Information_loss_ARD_Best}, we observe slightly higher FID scores for the ARD-VAE on the MNIST and CelebA dataset, indicating some loss of information relative to the baseline. However, the ARD-VAE outperforms the \emph{best} baseline on the challenging CIFAR10 and ImageNet datasets. Therefore, using extensive comparisons, we demonstrate the robustness and generalization capability of the ARD-VAE compared to the regular VAE and its variants.

\textbf{Relevant axes for the VAE using the Jacobian: } In this experiment we estimate the importance of the latent axes of a trained VAE using \ref{eq:ARD-VAE_rel_axes_estimation}.
\begin{align}\label{eq:ARD-VAE_rel_axes_estimation}
    &\w{}_{\boldsymbol{\hat{\sigma}}} = \sum_{k=1}^{D}\frac{1}{N}\sum_{i=1}^{N}\|\mathbb{J}_{i}\|^2, \\ \nonumber
    \text{ where } \mathbb{J} &= \left[ \dfrac{\partial \hat{\x{}}}{\partial \mu_{1}} \cdots \dfrac{\partial \hat{\x{}}}{\partial \mu_{L}} \right] \in \R^{D \times L} \text{ is the Jacobian matrix}.
\end{align}
To get a reliable estimate of the weight vector $\w{}_{\boldsymbol{\hat{\sigma}}}$ using \ref{eq:ARD-VAE_rel_axes_estimation}, we compute the Jacobian for multiple data samples, which is typically the size of a minibatch ($100$) in this work.

This estimate is used to determine the relevant axes for the VAE, $\beta$-TCVAE, RAE, and WAE and the selected axes are used for model evaluations as shown in Table \ref{tab:FID_Score_complete_Rel_Axes}. The relevant dimensions of the VAE and $\beta$-TCVAE are less than the initial dimension $L$ for the MNIST and CelebA dataset with comparable estimates of the FID scores. However, similar to the GECO-$L_0$-ARM-VAE and MaskAAE, majority of the dimensions collapses for the complex CIFAR dataset. This experiment demonstrates the robustness of the proposed ARD-VAE across different evaluation scenarios. There is no change in the number of active latent dimensions from $L$ for the RAE and WAE as both methods match aggregate posterior distributions, unlike VAEs.

\begin{table}
    \centering
    \resizebox{0.4\textwidth}{!}
    {
    \begin{tabular}{c c c c c}
    \hline
    Method &$L$ &$2L$ &$4L$ \\ [0.5ex] 
    \hline
    RAE $(L=128)$ &$104.90$ &$116.94$ &$121.50$ \\
    ARD-VAE $(L=128)$ &$85.25$ &$84.55$ &$84.60$ \\
    [1.1ex] \hline
    \end{tabular}
    }
    \caption{The FID scores of the RAE (second best method on the CIFAR10 dataset) and ARD-VAE on the CIFAR10 dataset with samples generated using all the latent dimensions. The FID scores of the RAE increase with the increase in the size of the latent space. Whereas, the FID scores of the ARD-VAE is consistent.} \label{tab:RAE_ARD-VAE_CIFAR10}
\end{table}

\begin{table*}
    \centering
    \resizebox{\textwidth}{!}
    {
    \begin{tabular}{c c c c c c c c c c c}
    \hline
    \multirow{2}{*}{\textsc{Bottleneck size}}
    &&\multicolumn{4}{c}{CIFAR10 $(L=128)$}
    &&\multicolumn{4}{c}{ImageNet $(L=256)$} \\
    \cline{3-6}\cline{8-11}
    &&\multirow{2}{*}{\textsc{Initial}} &\multirow{2}{*}{\textsc{Active}} &\textsc{Pruned Axes} &\multirow{2}{*}{\textsc{FID}$\downarrow$}
    &&\multirow{2}{*}{\textsc{Initial}} &\multirow{2}{*}{\textsc{Active}} &\textsc{Pruned Axes} &\multirow{2}{*}{\textsc{FID}$\downarrow$}\\ 
    &&&&$= \textsc{Initial}-$\textsc{Active} &
    &&&&$= \textsc{Initial}-$\textsc{Active} & \\
    \hline
    $L/2$
    &&$64$ &$60.40 \pm 0.49$ &\color{blue}{3.60} & ${104.03 \pm 1.33}$
    &&$128$ &$101.33 \pm 0.47$ &\color{blue}{26.67} & $129.63 \pm 1.26$ \\
    $L$
    &&$128$ &$105.80 \pm 1.33$ &\color{blue}{22.20}  &${87.56 \pm 1.21}$
    &&$256$ &$152.00 \pm 1.10$ &\color{blue}{104.00} & ${121.21 \pm 1.16}$ \\
    [1.2ex] \hline
    \end{tabular}
    }
    \caption{In this result, we address the issue when the initial size of latent space $L$ is smaller than the optimum size identified by the ARD-VAE. In this analysis, we study the FID scores of the generated samples and the number of \text{Active} dimensions identified with the reduced size of the latent space for the complex CIFAR10 and ImageNet datasets. The ARD-VAE \emph{preserves} most of the latent dimensions for the reduced size of the latent space, i.e., $L/2$, relative to the bigger latent space, $L$, which shows the effectiveness of the proposed method. As expected, the number of \textsc{Active} dimensions is less under the setting of $L/2$, subsequently increasing the FID scores.}
    \label{tab:ARD_less_than_optimum}
\end{table*}

\textbf{Size of the latent space for modeling a new dataset: } We know, the size of the latent space of the VAE and its variants for a dataset and a given autoencoder architecture is determined using cross-validation on the reconstruction loss. Therefore, we have to retrain the DLVM multiple times on a \emph{new} dataset with additional hyperparameter tuning to determine the number of latent dimensions, as the complexity of the dataset is \emph{unknown} to us. This is an unrealistic and tedious approach. In contrast, the ARD-VAE can start training with a reasonable estimate of $\beta \in [0.05, 0.5]$ (refer to \cref{tab:hyper-parameters} for the real datasets) and sufficiently large latent space to learn meaningful representations and model the data distribution.

In this experiment, we demonstrate the impact of an incorrectly chosen latent dimension, i.e., significantly bigger, on the performance of the RAE\cite{RAE} that produces comparable results to the ARD-VAE under different evaluation scenarios studied in this work. From the results reported in \cref{tab:RAE_ARD-VAE_CIFAR10}, we observe that the performance of the RAE is strongly affected by the choice of the initial size of the latent space. In contrast, the ARD-VAE is not sensitive to the initial size of the latent space and produces consistent results with $\beta= 0.05$. The results in \cref{tab:RAE_ARD-VAE_CIFAR10} illustrate the necessity of methods, such as the ARD-VAE, that can automatically identify the relevant latent dimensions required to model the distribution of a real dataset.

In another experimental setting, we adopt a conservative approach and train the ARD-VAE with a reduced number of latent dimensions on the CIAFR10 and ImageNet datasets than the optimum dimensions determined by the ARD-VAE. For example, we train the ARD-VAE on the CIFAR10 dataset in a latent space of size $L/2=64$ when the optimum size determined  determined by the ARD-VAE is $\sim106$. The motivation of this experiment is to evaluate the impact on the metric scores produced and the number of relevant dimensions identified by the ARD-VAE under such settings. We keep the hyperparameter \emph{unchanged} when using the reduced size of the latent space, such as $L/2$. From the results reported in the \cref{tab:ARD_less_than_optimum}, we observe that the ARD-VAE prunes less dimensions when trained on a smaller latent space and as expected, the FID scores increase. However, the impact on the metric scores is not significant.
